\pdfoutput=1

\documentclass[11pt]{article}

\usepackage{ACL2023}

\usepackage{times}
\usepackage{latexsym}

\usepackage[T1]{fontenc}

\usepackage[utf8]{inputenc}

\usepackage{microtype}

\usepackage{inconsolata}
\usepackage{graphicx}
\usepackage{subfigure}
\usepackage{amsmath}
\usepackage{amssymb}
\usepackage{booktabs}
\usepackage{stfloats}
\usepackage{float}
\graphicspath{{figure/}}
\usepackage{multirow}

\usepackage{algorithm}
\usepackage{algorithmic}
\usepackage{bbding}

\usepackage{xcolor}
\usepackage{pifont}

\title{Look Before You Leap: Towards Decision-Aware and Generalizable Tool-Usage for Large Language Models}

\author{
    Anchun Gui$^{\spadesuit\clubsuit}$\thanks{~~Equal contribution. Work was done during Anchun's internship at Tencent AI Lab.}\quad
    Jian Li$^{\clubsuit\ast}$\quad
    Yong Dai$^{\clubsuit}$\quad
    Nan Du$^{\clubsuit}$\quad
    Han Xiao$^{\spadesuit}$\thanks{~~Corresponding author.} \\
    $^{\spadesuit}$Xiamen University\quad
    $^{\clubsuit}$Tencent AI Lab \\
    \texttt{anchungui@stu.xmu.edu.cn\quad
    jackjianli@tencent.com}
}

\begin{document}

\maketitle

\begin{abstract}
Tool-augmented large language models (LLMs) are attracting widespread attention when accessing up-to-date knowledge and alleviating hallucination issues. Nowadays, advanced closed-source LLMs (e.g., ChatGPT) have demonstrated surprising tool-usage capabilities through prompting and in-context learning techniques. To empower the capabilities of open-source LLMs (e.g., LLaMA) in manipulating tools, current efforts focus on either template-driven or token-triggered tool-usage. However, the former hampers LLMs' flexibility to address diverse user's queries due to constrained tool interactions, while the latter limits the generalizability when engaging with new tools, since tool-usage learning is based on task- and tool-specific datasets. To alleviate these concerns, in this paper, we propose a \underline{de}cision-aware and gen\underline{er}alizable tool-usage framework (DEER). Specifically, we first construct the tool-usage samples with multiple decision branches via an automatic generation pipeline, thereby inspiring the decision-making awareness of LLMs under diverse scenarios. Meanwhile, we propose a novel tool sampling strategy to enhance the generalizability of LLMs over unseen tools. Extensive experiments demonstrate that our proposed DEER is effective and significantly outperforms baselines across various datasets. The source codes are available at \url{https://github.com/Ericmututu/ToolDEER}.
\end{abstract}

\section{Introduction}
\label{sec:intro}
Despite tremendous advancements in large language models (LLMs) \cite{touvron2023llama, touvron2023llamaa, openai2023gpt4}, the issues of hallucination \cite{ji2022survey} and unaccessible up-to-date knowledge limit LLMs to address professional and real-time queries. Tool-augmented LLMs, as a promising solution, are receiving extensive interest \cite{mialon2023augmented, qin2023tool}.
On one hand, tool-usage allows LLMs to access domain-specific knowledge, which is beneficial in alleviating hallucination issues.
On the other hand, the use of tools enables LLMs to interact with external real-world, which provides more potential applications for LLMs, such as weather querying, hotel booking, online shopping, etc.
Although some advanced closed-source LLMs (e.g., ChatGPT, GPT-4 \cite{openai2023gpt4}) have demonstrated surprising tool-usage capabilities by leveraging the prompting and in-context learning techniques, there is still a significant gap in compact open-source LLMs (e.g., LLaMA \cite{touvron2023llama}, Alpaca \cite{alpaca}).

\begin{figure}[!t]
    \centering
    \includegraphics[width=0.5\textwidth]{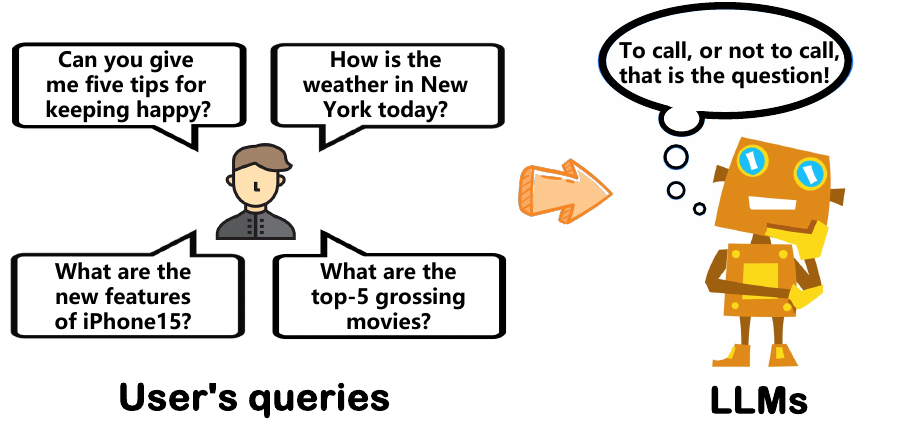}
    \caption{Given diverse user's queries, LLMs are expected to make optimal decisions for diverse queries to reduce unnecessary tool-usage and accelerate inference.}
    \label{fig:intro}
\end{figure}

\begin{figure*}[!t]
    \centering
    \includegraphics[width=1.0\textwidth]{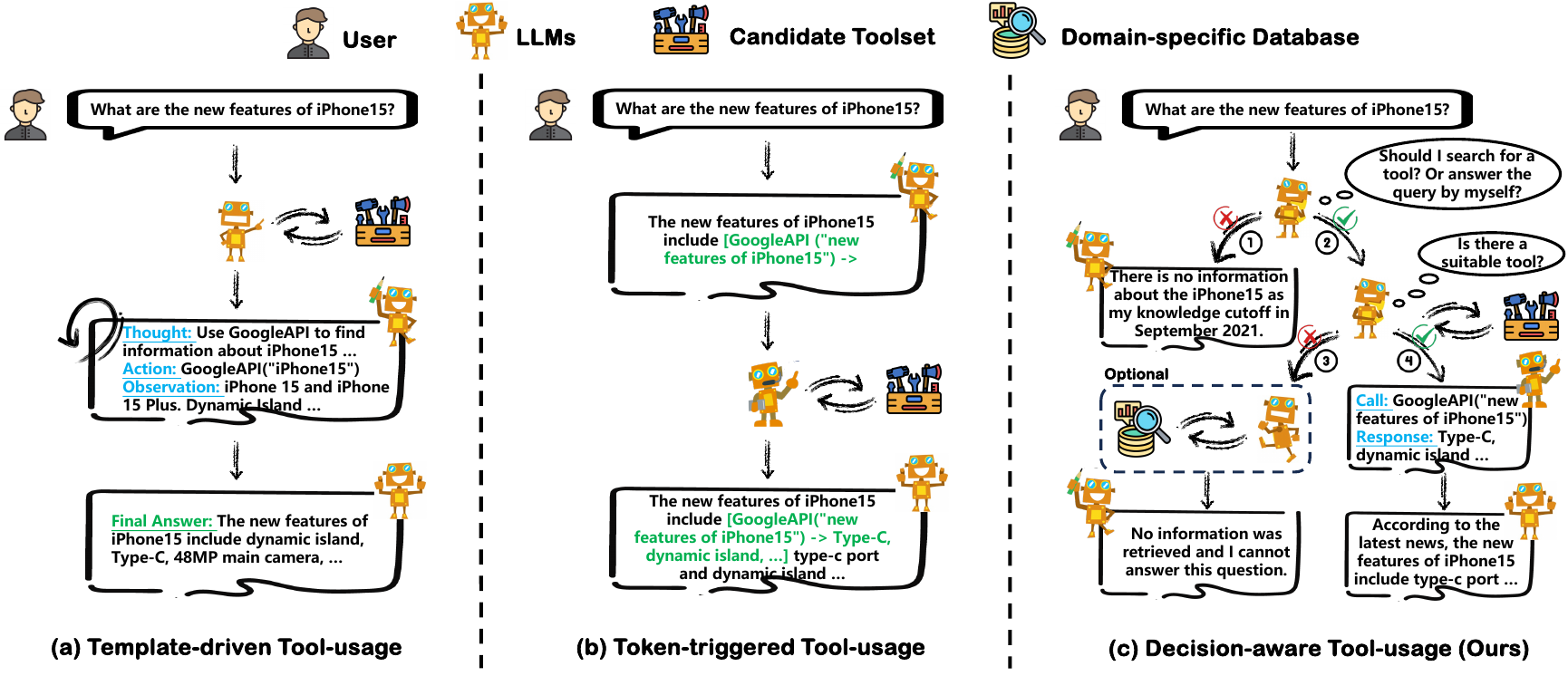}
    \caption{Comparison of different tool-usage paradigms.
    (a) In template-driven tool-usage, for any query, the interaction with tools is continuously implemented following a specific format (e.g., ``\textcolor{cyan}{\texttt{Thought-Action-Observation}}'') until obtaining the final answer.
    (b) In token-triggered tool-usage, the tool can be triggered by generating a specific token (e.g., ``\textcolor{cyan!20!green}{->}'') during inference.
    (c) In our decision-aware tool-usage, we design multiple decision branches (i.e., \ding{172}, \ding{173}, \ding{174}, \ding{175}) to address diverse user's queries, deciding whether should search for tools and whether there are suitable tools.
    Note that the candidate toolset indicates the set of currently provided tools for the model.}
    \label{fig:compare}
\end{figure*}

Recent research on the tool-usage of open-source LLMs can be generally categorized into two paradigms, as illustrated in Figure~\ref{fig:compare}:
(\romannumeral 1) Template-driven tool-usage.
First, the available tools (including their names, descriptions, APIs) are provided in the system prompt.
Then, for any user's query, the model is constrained to interact with tools following a specific format (e.g., ``\texttt{Thought-Action-Observation}'').
The final answer is obtained when reaching the maximum rounds of interaction or outputting a special terminator.
Representative works include ToolLLaMA \cite{qin2023toolllm}, ToolAlpaca \cite{tang2023toolalpaca}, etc.
(\romannumeral 2) Token-triggered tool-usage.
During inference, the decoding process is paused when encountering a particular token (e.g., ``\texttt{->}'').
Then, the calling command is extracted to request the corresponding API.
Next, the API's response is appended to the currently generated text before the decoding process resumes.
Representative works include TALM \cite{parisi2022talm}, Toolformer \cite{schick2023toolformer}, ToolkenGPT \cite{hao2023toolkengpt}, etc.

\textbf{Challenges \& Motivations.} Although these methods achieve the interaction between LLMs and tools, there are still some limitations:
(\romannumeral 1) Lack of decision-making awareness for tool-usage.
For template-driven tool-usage, the interaction process must follow a specific format for any user's query, which causes LLMs to lose flexibility when handling diverse queries.
In fact, an ideal scenario in Figure~\ref{fig:intro} is that, for general queries (e.g., ``Can you give me five tips for keeping happy?''), LLMs should give an answer with their own knowledge rather than resorting to external tools.
Besides, we expect that LLMs could stop tool calls when there is no suitable API in available tools, in case incurring invalid tool-usage and sluggish inference.
(\romannumeral 2) Lack of generalization on unseen tools.
For token-triggered tool-usage, such as Toolformer, which is trained on task- and tool-specific datasets, we cannot transfer it directly to the new tools.

\begin{figure*}[!t]
    \centering
    \includegraphics[width=0.9\textwidth]{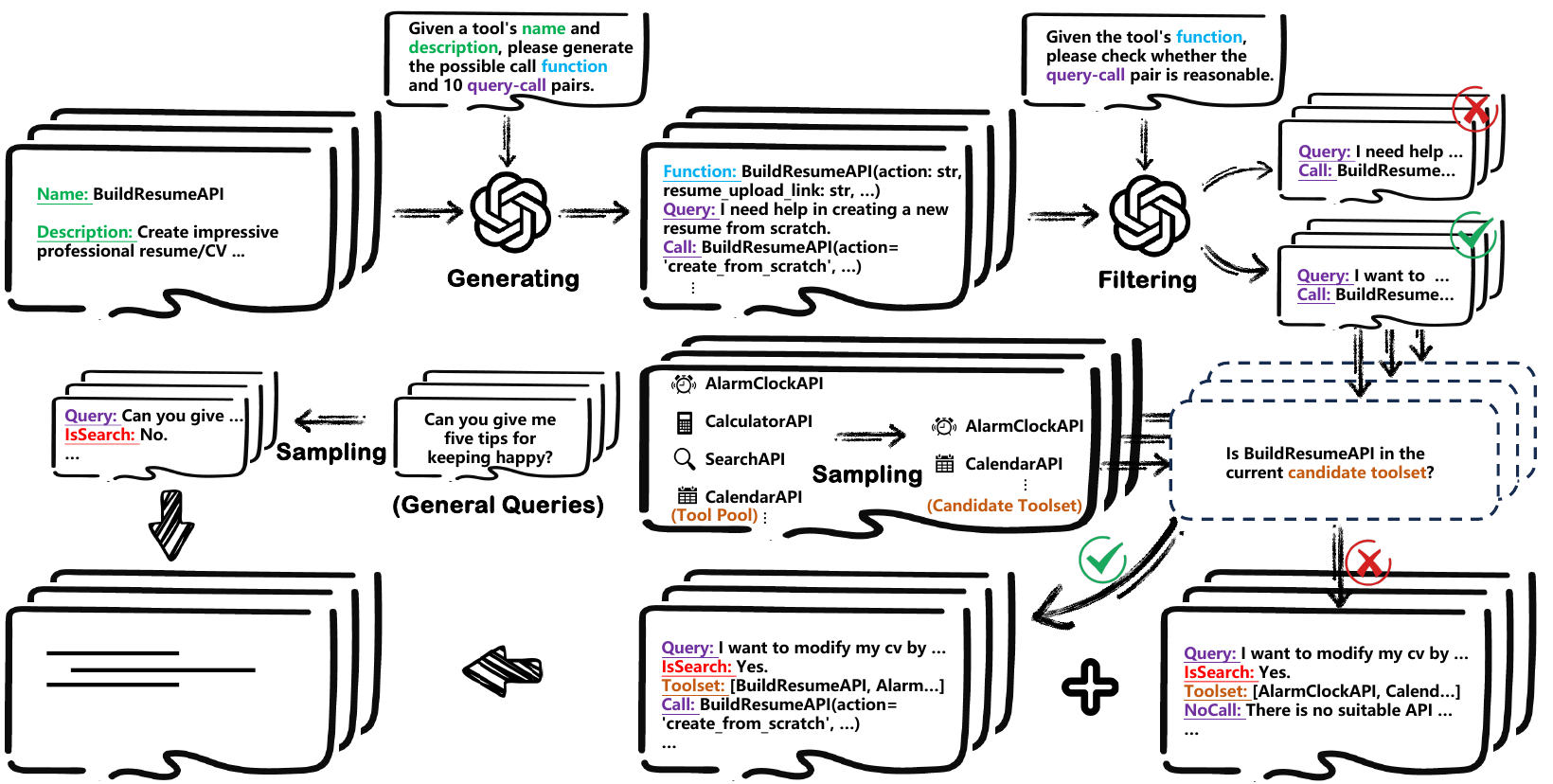}
    \caption{Pipeline of our sample generation. Detailed descriptions are presented in Section~\ref{subsec:multi_decision}. Note that here we simplify the prompt templates and contexts for clarity, and full prompt templates are provided in Appendix~\ref{app:template}.}
    \label{fig:pipe}
\end{figure*}

To address these limitations, in this paper, we propose a novel \underline{de}cision-aware and gen\underline{er}alizable tool-usage framework (DEER) based on open-source LLMs.
For the first limitation, we devise multiple decision branches based on the following questions, e.g., ``Should I search for a tool? Or answer the query by myself?'', ``Is there a suitable tool?'', as shown in Figure~\ref{fig:compare}(c).
Then, we construct the tool-usage dataset under diverse branches via an automatic generation pipeline (see Figure~\ref{fig:pipe}).
To bolster the decision-making prowess, we employ supervised fine-tuning (SFT) to train the model on our designated dataset.
For the second limitation, we propose a mixture of sampling strategies to boost the generalization of LLMs on unseen tools, where \texttt{Inter-class}, \texttt{Intra-class}, and \texttt{Random} sampling strategies (see Section~\ref{subsec:tool_sampling} \& \ref{subsec:effect_sampling}) strengthen the diversity, consistency, and randomness of the available tools, respectively.

To verify the effectiveness of our proposed method, we perform extensive experiments on the tool-usage datasets. The results demonstrate that our models achieve state-of-the-art performance against other baselines. For instance, our model obtains $98.6\%$ and $88.2\%$ overall accuracy under the decisions of searching and calling, respectively, which significantly outperforms GPT-4 ($78.1\%$, $87.6\%$). Moreover, our models also achieve superior generalization on various datasets (with the unseen tools) compared to other counterparts.

To summarize, our contributions are as follows:
\begin{itemize}
    \item We propose a novel decision-aware and generalizable tool-usage framework (DEER) to enhance the decision-making awareness of LLMs when addressing diverse queries.
    \item We investigate the effect of diverse tool sampling strategies on model performance, and propose a mixture strategy to improve the generalizability of LLMs on new tools.
    \item Extensive experiments demonstrate the effectiveness of our proposed method, and our model obtains state-of-the-art performance in diverse scenarios compared to other baselines.
\end{itemize}

\section{Related Work}
\subsection{Tool-Augmented Language Models}
Recently, several studies have demonstrated that closed-source LLMs (e.g., ChatGPT, GPT-4 \cite{openai2023gpt4}) have surprising tool-usage capabilities via prompting and in-context learning techniques \cite{wei2022chainofthought, press2023measuring, hsieh2023tool}. For example, ReAct \cite{yao2023react} synergies the reasoning and acting processes using LLMs.
Similar works include Chameleon \cite{lu2023chameleon}, HuggingGPT \cite{shen2023hugginggpt}, RestGPT \cite{song2023restgpt}, GPT4Tools \cite{yang2023gpt4tools}, and ART \cite{paranjape2023art}, which are based on the top of closed-source LLMs.
In addition, some works propose benchmarks (e.g., API-Bank \cite{li2023apibank}, ToolQA \cite{zhuang2023toolqa}, MetaTool \cite{huang2023metatool}) to evaluate the effectiveness of LLMs' tool-usage.
Given the comprehensive investigation of closed-source LLMs, this paper primarily emphasizes enhancing the tool-usage capabilities of open-source LLMs (e.g., LLaMA \cite{touvron2023llama}, Alpaca \cite{alpaca}, Vicuna \cite{zheng2023judging}).

\subsection{Tool-Usage with Open-Source LLMs}
Currently, based on compact open-source LLMs, some works have been proposed under the template-driven paradigm, e.g., 
ToolLLaMA \cite{qin2023toolllm}, ToolAlpaca \cite{tang2023toolalpaca}, Gorilla \cite{patil2023gorilla}, and \citet{xu2023tool}.
Meanwhile, several studies are included in the token-triggered paradigm such as TALM \cite{parisi2022talm}, Toolformer \cite{schick2023toolformer}, ToolkenGPT \cite{hao2023toolkengpt}.
Although these methods allow LLMs to interact with external tools, they neglect the decision-making awareness and generalization of LLMs in diverse scenarios.
For this reason, in this paper, we propose multi-decision branches of tool-usage and a novel tool sampling strategy to address these concerns.

\section{Methodology}
\subsection{Multi-Decision Branches Design}
\label{subsec:multi_decision}
In contrast to previous methods that do not consider query difficulty and tool suitability, we expect the model to ``Look Before You Leap'' in different scenarios, as illustrated in Figure~\ref{fig:compare}(c).
We achieve this goal by constructing a customized dataset with diverse branches and training our model on it.
Specifically, for each query, the model is asked the first question ``Should I search for a tool? Or answer the query by myself?'' (denoted as \texttt{Decision-Search}).
There are two cases:
\ding{172} the query can be directly answered by the model without searching for tools;
\ding{173} the query is beyond the model's knowledge, then the currently available tools (denoted as \texttt{Candidate Toolset}) are provided to the model.
Subsequently, based on case \ding{173}, the model is asked the second question ``Is there a suitable tool?'' (denoted as \texttt{Decision-Call}).
There are also two scenarios:
\ding{174} the answer is no, where the tool-usage process is terminated then the query is still answered by the model itself (note that we provide an optional retrieval process for those domain-specific queries if the database has been constructed);
\ding{175} the answer is yes, then the interaction between LLMs and tools is indeed engaged, where the model produces the API call command then receives the corresponding response\footnote{In fact, this process can be repeated several times, similar to the template-driven method. However, the difference in our setting is that, the model requires to determine the necessity of tool-usage before indeed implementing.}.

\paragraph{Automatic Generation Pipeline.}
Based on the above considerations, we construct the tool-usage samples under diverse scenarios via an automatic generation pipeline as shown in Figure~\ref{fig:pipe}.
To begin with, we provide the tool's name and description in the prompt and ask GPT-4 to generate plausible function and $10$ query-call pairs.
Considering the randomness of generation, we filter out unreasonable pairs by leveraging GPT-4 again with another specific prompt.
Then, for each query, we build a candidate toolset by sampling several tools from the set of all tools (i.e., the tool pool).
There are two possible scenarios:
(\romannumeral 1) the optimal tool for the current query lies in the candidate toolset, then we fill the corresponding command of calling API in the prompt, denoted as \texttt{Call} sample;
(\romannumeral 2) otherwise, we append the instruction that there is no suitable tool, denoted as \texttt{NoCall} sample.
Next, we sample some queries from a general conversational dataset for examples that do not need to search for tools (denoted as \texttt{NoSearch} sample).
Finally, our dataset consists of \texttt{NoSearch}, \texttt{NoCall}, and \texttt{Call} samples.

\begin{figure}[!t]
    \centering
    \includegraphics[width=0.45\textwidth]{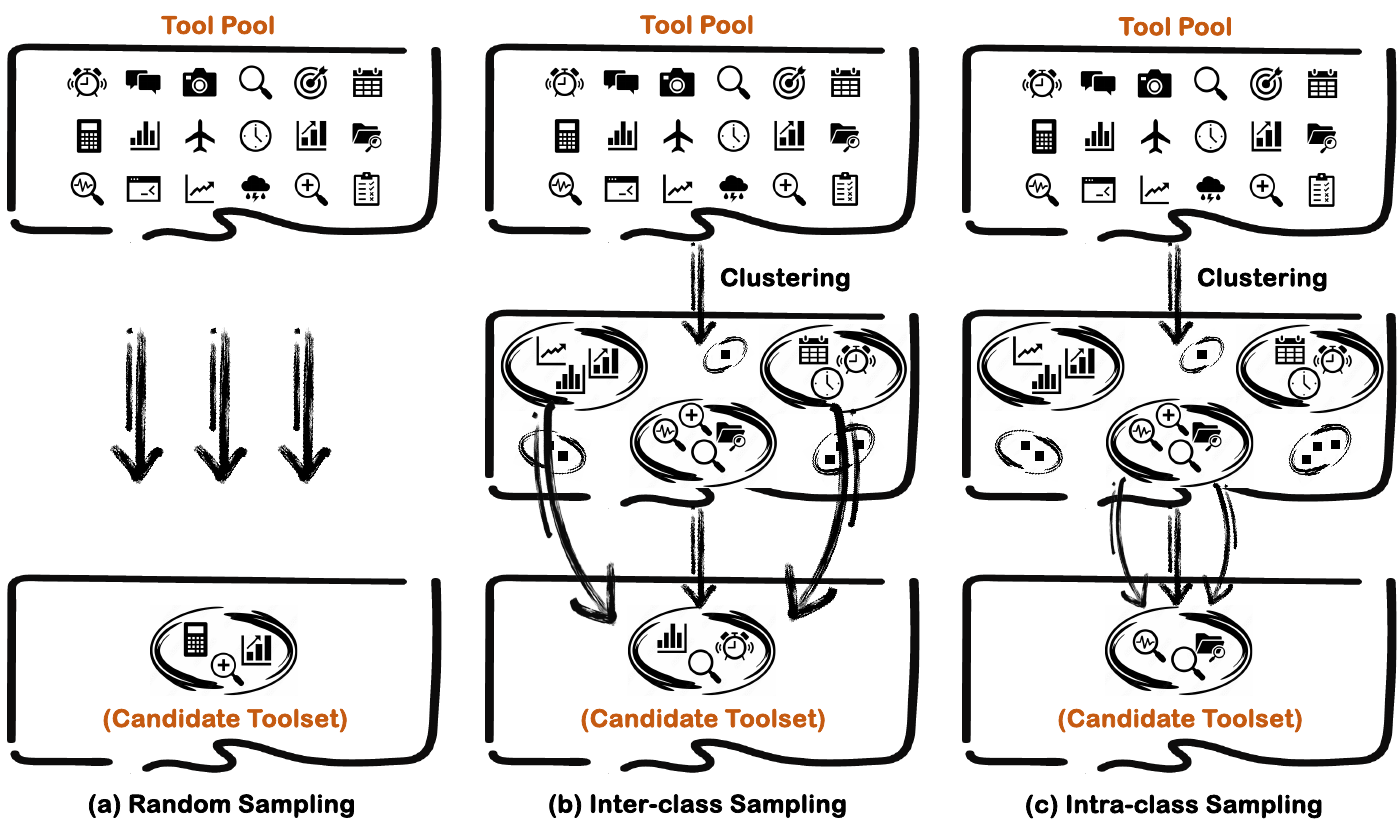}
    \caption{Illustration of different sampling strategies.}
    \label{fig:sampling}
\end{figure}

\subsection{Tool Sampling}
\label{subsec:tool_sampling}
The reason why we need to construct the candidate toolset is that, limited by the maximum input length of LLMs, it is impractical to incorporate all tool information into the prompt at once. Furthermore, in real-world applications, new tools may become available and be required after model training and deployment. Therefore, we expect the model to have a good generalization capability when the unseen tools lie in the candidate toolset.

To this end, we propose to train our model under randomly sampled candidate toolsets with or without ground-truth tools, which is similar to the ``Dropout'' strategy during training DNNs \cite{srivastava2014dropout}.
Specifically, we assume the set of all tools $\mathcal{T} = \{t_{i}\}_{i=1}^{N}$, where ${N}$ is the number of tools and each tool ($t$) consists of its name ($n$), description ($d$), and function ($f$), i.e., $t = \{n, d, f\}$.
For the query ($q$) that invokes the tool ($t_{q}$), the candidate toolset ($\mathcal{T}_{c}$) is constructed in the following three ways:
(\romannumeral 1) \texttt{Random}: we randomly take $k$ tools from $\mathcal{T}$ as $\mathcal{T}_{c}$, i.e., $\mathcal{T}_{c} = \{t_{i} | t_{i} \in \mathcal{T} \}_{i=1}^{k}$.
(\romannumeral 2) \texttt{Inter-class}: we first obtain the vectorized representation of tool $e(t)$ by embedding the tool's description $d$, i.e., $e(t) = \mathrm{embedding}(t(d))$\footnote{In our experiments, we take the SentenceTransformer \cite{reimers-2020-multilingual-sentence-bert} as the embedding model.}. 
Then, leveraging the K-means algorithm, we categorize all tools into $m$ clusters, where the whole cluster set $\Omega = \bigcup_{i=1}^{m} C_{i}$ and each cluster $C_{i} = \{t^{(i)}_{1}, t^{(i)}_{2}, \cdots, t^{(i)}_{|c_{i}|} \}$ and $|c_{i}| \ge k$.
Last, the candidate toolset is formed by $\mathcal{T}_{c} = \{t | t \in C_{i}, C_{i} \in \Omega \}_{i=1}^{k}$, where we first randomly select $k$ clusters $C_{i} \in \Omega$, and then randomly select one tool $t$ in each $C_{i}$.
(\romannumeral 3) \texttt{Intra-class}: in contrast to \texttt{Inter-class} sampling, $\mathcal{T}_{c}$ is constructed as follows:
$\mathcal{T}_{c} = \{t_{i} | t_{i} \in C^{*}, C^{*} \in \Omega \}_{i=1}^{k}$, where we select one cluster $C^{*}$ that includes the query needed tool $t_{q}$, and $t_{i}$ is randomly sampled from $C^{*}$.
In Section~\ref{subsec:effect_sampling}, we investigate the effectiveness of each sampling strategy, and find that the optimal performance can be obtained by mixing \texttt{Random}, \texttt{Inter-class}, and \texttt{Intra-class} in a certain proportion (see Figure~\ref{fig:sampling_comp}).
The overall comparison of these three sampling strategies is illustrated in Figure~\ref{fig:sampling}, and the detailed algorithmic process is described in Appendix~\ref{app:implement}.

\begin{table*}[!h]
    \centering
    \scalebox{1.0}{
        \begin{tabular}{lllllll} 
            \toprule
                \multirow{2}{*}{Method} & \multicolumn{3}{c}{Decision-Search} & \multicolumn{3}{c}{Decision-Call} \\
            \cmidrule(l){2-4} \cmidrule(l){5-7} & $\rm{P}_{NoSearch}$ & $\rm{P}_{Search}$ & $\rm{P}_{DS}$ & $\rm{P}_{NoCall}$ & $\rm{P}_{Call}$ & $\rm{P}_{DC}$ \\
            \midrule
                LLaMA2-13B-Chat (2-shot) & $55.3_{\pm 11.6}$ & $66.5_{\pm 8.5}$ & $64.0_{\pm 6.4}$ & $54.5_{\pm 11.9}$ & $53.9_{\pm 3.9}$ & $54.2_{\pm 2.4}$ \\
                LLaMA2-13B-Chat (4-shot) & $46.9_{\pm 7.6}$ & $73.7_{\pm 6.6}$ & $68.8_{\pm 6.7}$ & $50.1_{\pm 11.5}$ & $55.7_{\pm 7.4}$ & $52.0_{\pm 8.4}$ \\
                LLaMA2-13B-Chat (6-shot) & $44.4_{\pm 9.1}$ & $74.4_{\pm 12.9}$ & $70.0_{\pm 7.5}$ & $73.5_{\pm 9.7}$ & $47.7_{\pm 9.8}$ & $56.9_{\pm 1.5}$ \\
            \midrule
                GPT-3.5-Turbo (2-shot) & $53.2_{\pm 7.8}$ & $63.3_{\pm 8.8}$ & $61.4_{\pm 6.1}$ & $69.0_{\pm 10.1}$ & $51.5_{\pm 9.1}$ & $58.5_{\pm 2.2}$ \\
                GPT-3.5-Turbo (4-shot) & $46.8_{\pm 5.8}$ & $68.3_{\pm 4.4}$ & $64.4_{\pm 2.8}$ & $61.2_{\pm 7.4}$ & $55.6_{\pm 10.5}$ & $57.9_{\pm 1.9}$ \\
                GPT-3.5-Turbo (6-shot) & $43.8_{\pm 6.5}$ & $72.5_{\pm 1.4}$ & $67.3_{\pm 0.9}$ & $59.2_{\pm 11.7}$ & $57.4_{\pm 6.9}$ & $58.2_{\pm 0.7}$ \\
            \midrule
                GPT-4 (2-shot) & $61.6_{\pm 9.8}$ & $78.4_{\pm 4.9}$ & $75.3_{\pm 2.8}$ & $87.1_{\pm 2.9}$ & $86.9_{\pm 1.4}$ & $86.9_{\pm 0.6}$ \\
                GPT-4 (4-shot) & $70.3_{\pm 10.1}$ & $77.3_{\pm 5.2}$ & $76.1_{\pm 2.5}$ & $\underline{87.4_{\pm 2.2}}$ & $87.2_{\pm 0.4}$ & $87.3_{\pm 0.8}$ \\
                GPT-4 (6-shot) & $74.3_{\pm 5.7}$ & $78.9_{\pm 3.7}$ & $78.1_{\pm 2.3}$ & $87.3_{\pm 1.4}$ & $\underline{87.7_{\pm 0.8}}$ & $\underline{87.6_{\pm 0.3}}$ \\
            \midrule
                ToolDEER-7B & $\underline{94.1_{\pm 1.6}}$ & $\underline{98.4_{\pm 2.4}}$ & $\underline{97.7_{\pm 2.1}}$ & $83.5_{\pm 2.7}$ & $83.7_{\pm 1.8}$ & $83.4_{\pm 1.9}$ \\
                ToolDEER-13B & $\textbf{95.9}_{\pm 1.9}$ & $\textbf{99.2}_{\pm 0.4}$ & $\textbf{98.6}_{\pm 0.3}$ & $\textbf{88.9}_{\pm 3.1}$ & $\textbf{88.4}_{\pm 2.3}$ & $\textbf{88.2}_{\pm 2.0}$ \\
            \bottomrule
        \end{tabular}
    }
    \caption{The comparison of model's decision-making awareness under \texttt{Decision-Search} and \texttt{Decision-Call} scenarios, respectively. Here, the baselines (i.e., LLaMA2-13B-Chat, GPT-3.5-Turbo, GPT-4) are equipped with $2$, $4$, $6$ demonstrations, respectively, within in-context learning. Each result is averaged over six trials, where the subscript is the standard deviation. The first and second best results are marked with \textbf{bold} and \underline{underline}, respectively.}
    \label{tab:main_exp}
\end{table*}

\begin{table}[!h]
    \centering
    \scalebox{0.95}{
        \begin{tabular}{lcccc}
            \toprule
                \multicolumn{1}{c}{\multirow{2}{*}{}} & \multirow{2}{*}{\#NoSearch} & \multicolumn{2}{c}{\#Search} & \multicolumn{1}{c}{\multirow{2}{*}{\#Total}} \\
                \cmidrule(l){3-4} \multicolumn{1}{c}{} & & \#NoCall & \#Call & \multicolumn{1}{c}{} \\
            \midrule
                Train & $1,807$ & $3,114$ & $4,664$ & $9,585$ \\
                Valid & $193$ & $346$ & $526$ & $1,065$ \\
                Test  & - & $298$ & $445$ & $743$ \\
            \bottomrule
        \end{tabular}
    }
    \caption{Statistics of our datasets. \# refers to the number of samples in the corresponding decision scenario.}
    \label{tab:stat}
\end{table}

\section{Experiments}
\subsection{Setup}
\label{subsec:setup}
\paragraph{Datasets \& Evaluation Protocols.}
Firstly, we collect $977$ tools (including their names and descriptions) by filtering ChatGPT-Plugins JSON\footnote{\url{https://github.com/copilot-us/chatgpt-plugins/tree/main}}.
Then, in \texttt{Decision-Call} scenario (i.e., \texttt{Search}), we construct a total of $8,650$ samples based on $900$ randomly selected tools, where the number of samples in \texttt{NoCall} and \texttt{Call} are $3,460$ and $5,190$, respectively.
For those \texttt{NoSearch} samples, we randomly select $2,000$ queries from general conversational dataset \cite{xu2023wizardlm}.
To evaluate the generalizability of models on unseen tools, we construct $743$ test samples based on $77$ additional tools (which are not involved in training), where the number of samples under \texttt{NoCall} and \texttt{Call} are $298$ and $445$, respectively.
Detailed statistics of our datasets are shown in Table~\ref{tab:stat}.
In addition to our test set, we also conduct experiments on ToolBench \cite{qin2023toolllm} and ToolCorpus \cite{tang2023toolalpaca}\footnote{For ToolBench, we choose G1-Category \cite{qin2023toolllm} as the test set; for ToolCorpus (we named), the test set consists of the real and simulated sets \cite{tang2023toolalpaca}.}.

To verify the effectiveness of our approach, we propose the following evaluation protocols:
(\romannumeral 1) In \texttt{Decision-Search} scenario, we define that
\begin{align}
    \scalebox{0.88}{$\rm{P}_{NoSearch} = \frac{n_{nos}}{N_{nos}}, \rm{P}_{Search} = \frac{n_{s}}{N_{s}}, \rm{P}_{DS} = \frac{n_{nos} + n_{s}}{N_{nos} + N_{s}}$}
\end{align}
where $\rm{n}_{nos}$, $\rm{n}_{s}$ refer to the number of correctly predicted samples in $\texttt{NoSearch}$, $\texttt{Search}$, respectively; $\rm{N}_{nos}$, $\rm{N}_{s}$ indicate the total number of samples in $\texttt{NoSearch}$, $\texttt{Search}$, respectively.
(\romannumeral 2) In \texttt{Decision-Call} scenario, we can define that
\begin{align}
    \scalebox{0.88}{$\rm{P}_{NoCall} = \frac{n_{noc}}{N_{noc}}, \rm{P}_{Call} = \frac{n_{c}}{N_{c}}, \rm{P}_{DC} = \frac{n_{noc} + n_{c}}{N_{noc} + N_{c}}$}
\end{align}
where $\rm{n}_{noc}$, $\rm{n}_{c}$ refer to the number of correctly predicted samples in $\texttt{NoCall}$, $\texttt{Call}$, respectively; $\rm{N}_{noc}$, $\rm{N}_{c}$ indicate the total number of samples in $\texttt{NoCall}$, $\texttt{Call}$, respectively. Here, $\rm{N}_{s} = \rm{N}_{noc} + \rm{N}_{c}$.

\begin{table*}[!h]
    \centering
    \scalebox{0.81}{
        \begin{tabular}{llllllllll} 
            \toprule
                \multirow{2}{*}{Method} & \multicolumn{3}{c}{Our Dataset (Test)} & \multicolumn{3}{c}{ToolBench \cite{qin2023toolllm}} & \multicolumn{3}{c}{ToolCorpus \cite{tang2023toolalpaca}} \\
            \cmidrule(l){2-4} \cmidrule(l){5-7} \cmidrule(l){8-10} & $\rm{P}_{NoCall}$ & $\rm{P}_{Call}$ & $\rm{P}_{DC}$ & $\rm{P}_{NoCall}$ & $\rm{P}_{Call}$ & $\rm{P}_{DC}$ & $\rm{P}_{NoCall}$ & $\rm{P}_{Call}$ & $\rm{P}_{DC}$ \\
            \midrule
                LLaMA2-13B-Chat$^{\dagger}$ 
                & $76.5_{\pm 11.7}$ & $48.1_{\pm 9.4}$ & $58.4_{\pm 3.7}$
                & $77.4_{\pm 5.4}$ & $33.3_{\pm 4.3}$ & $53.2_{\pm 5.6}$
                & $74.5_{\pm 9.7}$ & $72.0_{\pm 9.3}$ & $73.1_{\pm 5.7}$ \\
                GPT-3.5-Turbo$^{\dagger}$ 
                & $61.5_{\pm 4.7}$ & $59.6_{\pm 2.4}$ & $60.4_{\pm 1.3}$
                & $82.1_{\pm 5.8}$ & $64.9_{\pm 4.6}$ & $71.6_{\pm 2.3}$
                & $56.4_{\pm 12.8}$ & $86.9_{\pm 5.7}$ & $74.9_{\pm 6.3}$ \\
                GPT-4$^{\dagger}$ 
                & $\textbf{88.6}_{\pm 3.1}$ & $\underline{89.4_{\pm 1.5}}$ & $\underline{89.0_{\pm 1.2}}$
                & $\underline{86.5_{\pm 4.9}}$ & $\underline{84.7_{\pm 3.2}}$ & $\underline{85.9_{\pm 2.4}}$
                & $74.4_{\pm 8.3}$ & $\textbf{97.9}_{\pm 4.6}$ & $84.2_{\pm 5.4}$ \\
            \midrule
                ToolLLaMA-7B & $5.4_{\pm 1.7}$ & $79.3_{\pm 2.7}$ & $49.7_{\pm 2.8}$ & $24.7_{\pm 2.1}$ & $78.0_{\pm 0.9}$ & $55.0_{\pm 2.2}$ & $5.2_{\pm 1.7}$ & $91.8_{\pm 0.9}$ & $56.3_{\pm 0.7}$ \\
                ToolAlpaca-7B & $1.0_{\pm 0.9}$ & $84.1_{\pm 1.2}$ & $50.8_{\pm 0.8}$ & $28.8_{\pm 1.6}$ & $58.3_{\pm 0.4}$ & $46.5_{\pm 0.7}$ & $5.4_{\pm 2.3}$ & $90.7_{\pm 1.6}$ & $56.6_{\pm 1.2}$ \\
                ToolAlpaca-13B & $2.7_{\pm 1.7}$ & $90.3_{\pm 0.8}$ & $55.2_{\pm 0.9}$ & $16.3_{\pm 3.7}$ & $66.7_{\pm 3.1}$ & $46.5_{\pm 2.9}$ & $7.8_{\pm 1.5}$ & $93.9_{\pm 1.8}$ & $58.2_{\pm 1.4}$ \\
            \midrule
                ToolDEER-7B & $84.0_{\pm 1.9}$ & $85.2_{\pm 2.1}$ & $84.5_{\pm 1.6}$ & $85.4_{\pm 0.7}$ & $83.3_{\pm 1.2}$ & $84.5_{\pm 0.9}$ & $\underline{80.4_{\pm 2.1}}$ & $96.0_{\pm 1.0}$ & $\underline{89.7_{\pm 0.6}}$ \\
                ToolDEER-13B & $\underline{88.0_{\pm 2.3}}$ & $\textbf{91.5}_{\pm 1.5}$ & $\textbf{89.2}_{\pm 1.6}$ & $\textbf{88.5}_{\pm 1.4}$ & $\textbf{86.7}_{\pm 0.8}$ & $\textbf{87.0}_{\pm 1.3}$ & $\textbf{81.6}_{\pm 2.9}$ & $\underline{97.3_{\pm 2.3}}$ & $\textbf{90.5}_{\pm 1.0}$ \\
            \bottomrule
        \end{tabular}
    }
    \caption{The comparison of generalizability on unseen tools. The baselines$^{\dagger}$ are equipped with $6$ demonstrations within in-context learning. The first and second best results are marked with \textbf{bold} and \underline{underline}, respectively.}
    \label{tab:unseen}
\end{table*}

\paragraph{Baselines.}
Current advanced LLMs such as ChatGPT\footnote{Note that ChatGPT and GPT-3.5-Turbo are equivalent in our statements, unless otherwise specified.} and GPT-4 \cite{openai2023gpt4} are introduced as our strong baselines. To compare with the original model, we also introduce LLaMA2-13B-Chat \cite{touvron2023llamaa} as a baseline.
To evaluate their decision-making capabilities, we adopt the prompting and in-context learning techniques (where $2$, $4$, $6$ demonstrations are employed, respectively) to ask them to make optimal decision towards diverse queries.
Besides, we also introduce ToolLLaMA \cite{qin2023toolllm} and ToolAlpaca \cite{tang2023toolalpaca} as the counterparts when evaluating the generalizability on unseen tools.
In the experiments, our model is named as \textbf{ToolDEER}.

\paragraph{Implementation.}
In our experiments, LLaMA2-7B and LLaMA2-13B \cite{touvron2023llamaa} are leveraged as the backbone models. During training, we fine-tune these models for several epochs on the training set via supervised fine-tuning (SFT) \cite{ouyang2022training}, then we directly evaluate the fine-tuned models on the validation and test sets.
Note that the validation set is not involved in the training.
In addition, to preserve the model's original knowledge and general conversational capabilities as much as possible, we adopt LoRA-style \cite{hu2022lora} parameter-efficient fine-tuning. Finally, detailed configurations of hyperparameters (e.g., learning rate) are reported in Appendix~\ref{app:experiment}.

\subsection{Main Results}
\paragraph{The decision awareness of tool-usage.}
The results are reported in Table~\ref{tab:main_exp}, from which we can draw the following observations:
(\romannumeral 1) Regarding the first level of \texttt{Decision-Search}, although powerful LLMs show somewhat discriminative capability, for example, GPT-4 achieves $78.1\%$ accuracy with the help of $6$ demonstrations, they still fall short of a satisfactory performance.
From the comparison, our models achieve significant improvements after fine-tuning. For instance, ToolDEER-7B and ToolDEER-13B obtain $97.7\%$ and $98.6\%$ accuracy, respectively, in terms of determining whether a tool is needed for the query.
(\romannumeral 2) Regarding the second level of \texttt{Decision-Call}, for these baselines (LLaMA2-13B-Chat, GPT-3.5-Turbo), their best discriminative capabilities ($56.9\%$, $58.2\%$) are slightly better than random selection ($50\%$). In comparison, GPT-4 achieves a highly competitive result of $87.6\%$. This is understandable since our datasets are generated by GPT-4. Nonetheless, ToolDEER-13B still outperforms GPT-4 by a significant margin. Considering the small-scale and open-source properties, our models are quite appealing compared to closed-source LLMs.

\paragraph{The generalization of tool-usage on unseen tools.}
The results are shown in Table~\ref{tab:unseen}, where we can observe that:
(\romannumeral 1) ToolDEER-13B achieves state-of-the-art performance across various datasets, which demonstrates the superior generalization of our method on new tools. For example, on our dataset, ToolDEER-13B obtains $\rm{P}_{DC} = 89.2\%$ and even achieves $91.5\%$ accuracy in \texttt{Call} case. Moreover, based on ToolBench and ToolCorpus, ToolDEER-7B and ToolDEER-13B also outperform their counterparts with significant improvements, especially on ToolBench.
(\romannumeral 2) Our models not only maintain high generalization on unseen tools, but also keep high accuracy in \texttt{NoCall} scenario. In contrast, ToolLLaMA and ToolAlpaca have almost no such capability on our dataset and ToolCorpus. Note that on ToolBench, they have $16.3\%-28.8\%$ accuracy, which we speculate that this may be attributable to the disparate difficulty levels of these datasets.

\section{Analysis}

\begin{figure*}[!t]
    \centering
    \subfigure {
        \includegraphics[width=0.31\textwidth]{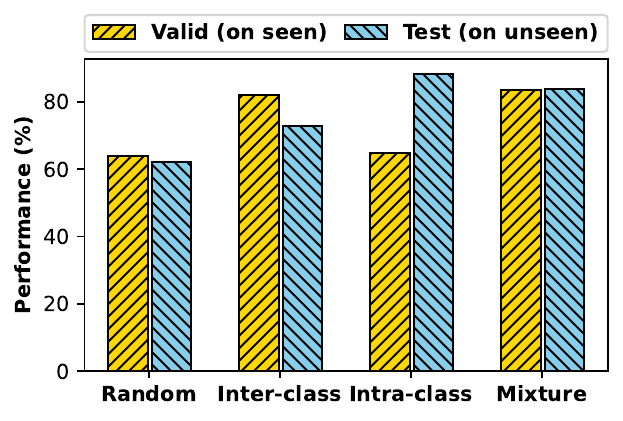}
    }
    \subfigure {
        \includegraphics[width=0.31\textwidth]{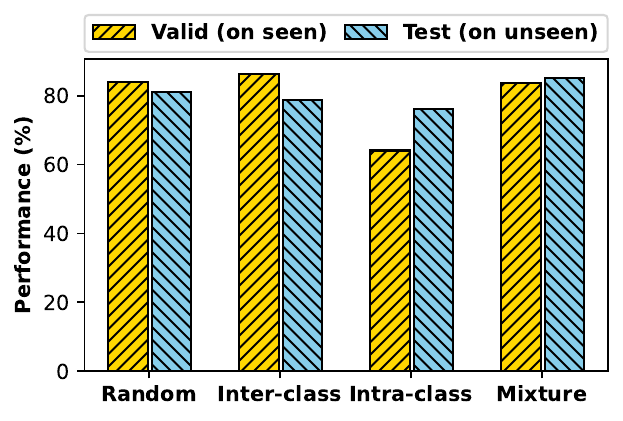}
    }
    \subfigure {
        \includegraphics[width=0.31\textwidth]{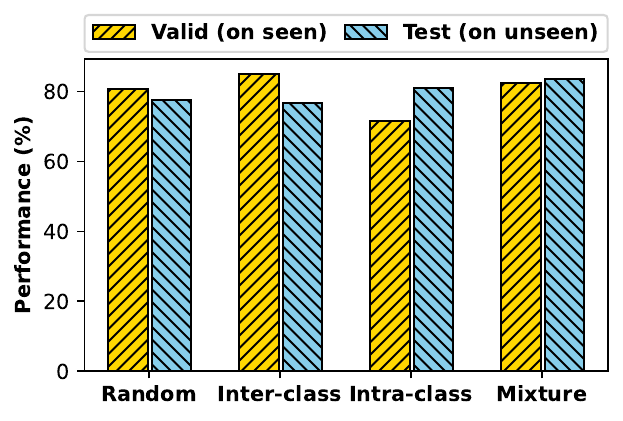}
    } \\
    \subfigure {
        \includegraphics[width=0.31\textwidth]{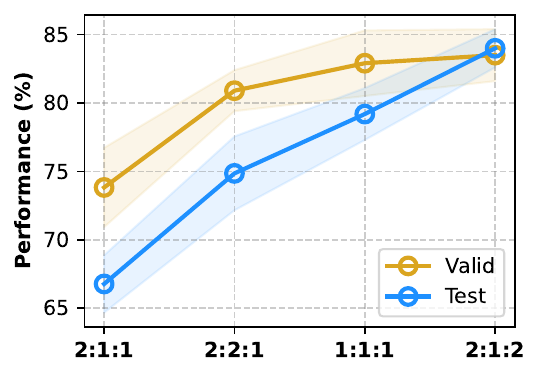}
    }
    \subfigure {
        \includegraphics[width=0.31\textwidth]{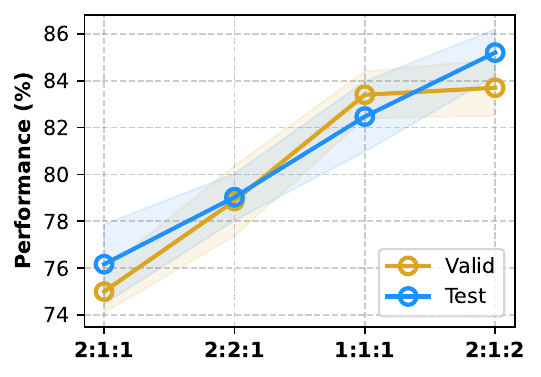}
    }
    \subfigure {
        \includegraphics[width=0.31\textwidth]{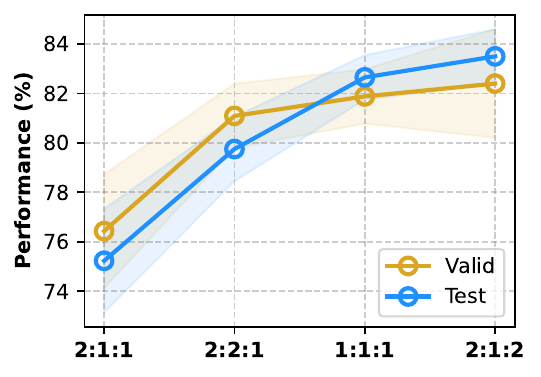}
    }
    \caption{Comparisons of diverse tool sampling strategies (in the first row) and sampling ratios (in the second row, where ``\texttt{Random} : \texttt{Intra-class} : \texttt{Inter-class}'' in the $x$-axis). Here, the experiments of the first, second, and third columns are conducted on the \texttt{NoCall}, \texttt{Call}, and \texttt{Decision-Call} scenarios (i.e., $\rm{P}_{NoCall}$, $\rm{P}_{Call}$, $\rm{P}_{DC}$), respectively. Note that the valid(ation) and test sets are built on seen and unseen tools, respectively.}
    \label{fig:sampling_comp}
\end{figure*}

\subsection{The Effect of Tool Sampling Strategy}
\label{subsec:effect_sampling}
In the main experiments, we achieve high accuracy and strong generalization on the validation and test sets, respectively, by mixing $\texttt{Random}$, $\texttt{Inter-class}$, and $\texttt{Intra-class}$ sampling strategies. Therefore, a natural research question is: How do different sampling strategies affect model performance?
To this end, we individually employ each sampling strategy during training and then explore its impact on the validation set (with seen tools) and test set (with unseen tools). 
Here, we use consistent experimental setups with Section~\ref{subsec:setup}.

The comparisons are reported in Figure~\ref{fig:sampling_comp}, where we can observe that:
(\romannumeral 1) Under $\texttt{Random}$ sampling, the model only obtains moderate accuracy and generalization on the validation and test sets.
(\romannumeral 2) Under $\texttt{Inter-class}$ sampling, the model achieves superior accuracy on the validation set, but the generalization on the test set is unsatisfactory.
(\romannumeral 3) Under $\texttt{Intra-class}$ sampling, while the accuracy is decreased on the validation set, surprisingly, the gains on the test set are significant.
Therefore, we can draw the following conclusions: $\texttt{Inter-class}$ sampling could enhance the accuracy but limits the generalization; in contrast, $\texttt{Intra-class}$ sampling decreases the accuracy yet significantly benefits the generalization; $\texttt{Random}$ makes a trade-off between accuracy and generalization.
This is understandable, here we provide some intuitive explanations from the perspective of contrastive learning.
For $\texttt{Inter-class}$ sampling, it enhances the diversity of candidate toolset by introducing ``negative'' tools from different clusters, which allows the model to identify the optimal tool in an easy mode. However, it also hinders the model from distinguishing among similar tools.
In contrast, $\texttt{Intra-class}$ sampling builds the candidate toolset by introducing ``positive'' tools from the cluster where the optimal tool is located, which results in highly similar tools in the current candidate toolset. Nevertheless, the model can still select the optimal tool through SFT training, which strongly boosts the model's understanding and discriminative capabilities for diverse tools. This may explain to some extent why the generalization on the test set is significantly improved under $\texttt{Intra-class}$ strategy.

Based on above observations, we propose to mix these three sampling strategies to achieve both high accuracy and strong generalization. The experimental results also support our assumption, referring to the \texttt{Mixture} sampling in Figure~\ref{fig:sampling_comp}.
In addition, we also explore the influence of sampling ratio in the second row of Figure~\ref{fig:sampling_comp}, where the state-of-the-art performance is obtained when $\texttt{Random}:\texttt{Intra-class}:\texttt{Inter-class}=2:1:2$.

\subsection{Ablation Study}
\begin{figure*}[!t]
    \centering
    \subfigure [\# NoSearch Samples] {
        \includegraphics[width=0.233\textwidth]{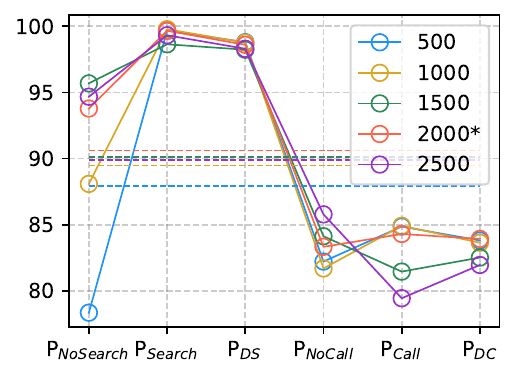}
        \label{subfig:abla_num_nosearch}
    }
    \subfigure [\# Tool Pool] {
        \includegraphics[width=0.233\textwidth]{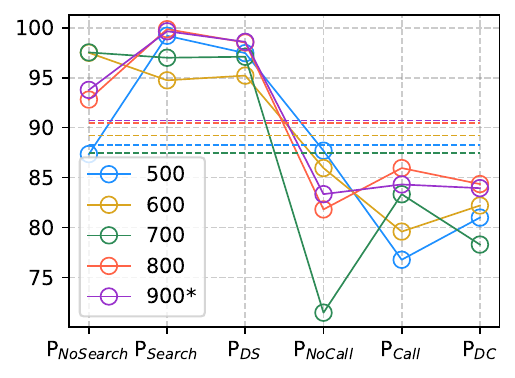}
        \label{subfig:abla_num_tool}
    }
    \subfigure [\# Candidate Toolset] {
        \includegraphics[width=0.233\textwidth]{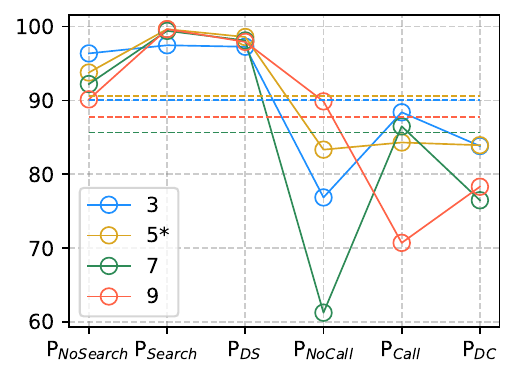}
        \label{subfig:abla_num_candidate}
    }
    \subfigure [\# Tool Clusters] {
        \includegraphics[width=0.233\textwidth]{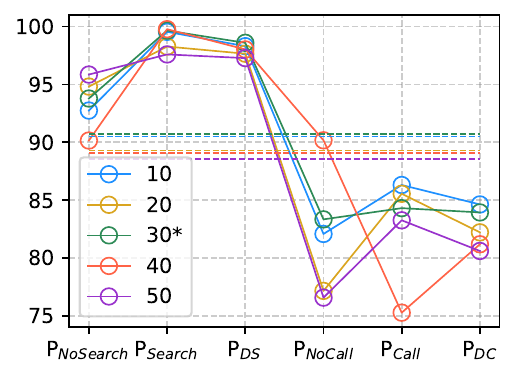}
        \label{subfig:abla_num_cluster}
    }
    \caption{Ablation experiments of hyperparameters, where the dashed line refers to the average of all metrics, \# indicates the number (or size) of corresponding item, and * denotes the setting we used in our experiments.}
    \label{fig:abla}
\end{figure*}

In this section, we mainly investigate the following research questions\footnote{The experiments in this section are performed on the validation set using LLaMA-7B as the backbone model. More experiments on hyperparameters are presented in Appendix~\ref{app:experiment}.}.
\textbf{RQ1:} How many samples in $\texttt{NoSearch}$ scenario can preserve the model's general conversational capability?
\textbf{RQ2:} How does the size of tool pool ($|\mathcal{T}|$) affect model performance?
\textbf{RQ3:} How does the number of tools in the candidate toolset ($|\mathcal{T}_{c}|$) affect the performance?
\textbf{RQ4:} In \texttt{Intra-} and \texttt{Inter-class} sampling, does the number of clusters ($|\Omega|$) affect the performance?
To answer these questions, we conduct extensive experiments under various settings in Figure~\ref{fig:abla}, from which we can observe that:
(\romannumeral 1) Overall, the performance is improved with the increasing of $\texttt{NoSearch}$ samples and $|\mathcal{T}|$.
(\romannumeral 2) The model is more likely to identify the most appropriate tool when the size of $|\mathcal{T}_{c}|$ and $|\Omega|$ are $5$ and $30$, respectively. And, more tool clusters may yield negative effects rather than further improving the performance.

\begin{table}[!t]
    \centering
    \scalebox{0.73}{
        \begin{tabular}{lcccc}
            \toprule
                Method & BLEU & ROUGE-1 & ROUGE-2 & ROUGE-L \\
            \midrule
                ToolLLaMA-7B  & $3.5$ & $26.0$ & $9.3$ & $16.6$ \\
                ToolAlpaca-7B & $5.7$ & $31.0$ & $12.7$ & $18.9$ \\
                ToolDEER-7B   & $\textbf{10.9}$ & $\textbf{42.7}$ & $\textbf{20.8}$ & $\textbf{26.5}$ \\
            \bottomrule
        \end{tabular}
    }
    \caption{The comparison with the original model on maintaining general conversational capacity.}
    \label{tab:general}
\end{table}

\subsection{Assessment of General Dialogue Capacity}
As we illustrated in Section~\ref{sec:intro}, previous paradigms such as template-driven tool-usage would limit LLMs' general conversational capabilities, since they are constrained to following specific generation templates to answer user's queries. To this end, in our tool-usage strategy, we reserve a quick decision path (i.e., $\texttt{NoSearch}$), where the query can be directly answered by LLMs without searching for the tool.
To demonstrate the effectiveness of our strategy, we provide quantitative comparisons with baselines in Table~\ref{tab:general}. Specifically, on $1,000$ queries randomly selected from the general conversational dataset, we take the output of original model as the ground-truth, and then measure the discrepancy between diverse model's output and ground-truth via BLEU and ROUGE-1/2/L (F1 score) metrics \cite{papineni2002bleu, lin2004rouge}.
As can be seen from Table~\ref{tab:general}, the outputs of our model have a high consistency with the original model compared to baselines, which suggests that our method greatly maintains the general dialog capability. More detailed examples are presented in Appendix~\ref{app:experiment}.

\begin{figure}[!t]
    \centering
    \includegraphics[width=0.5\textwidth]{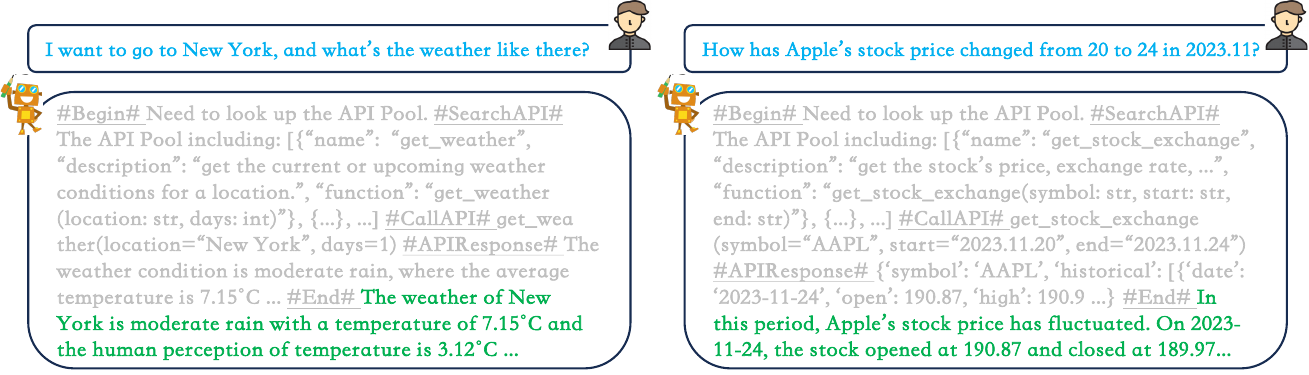}
    \caption{Examples of calling the real-world APIs (e.g., $\texttt{get\_weather}$, $\texttt{get\_stock\_exchange}$)\protect\footnotemark, where the \textcolor{gray}{gray} is an intermediate reasoning step (invisible for the user), the \textcolor{cyan!20!green}{green} is final response to the user's \textcolor{cyan}{query}.}
    \label{fig:case_study}
\end{figure}
\footnotetext{For the weather: \url{https://openweathermap.org/}; for the stock: \url{https://site.financialmodelingprep.com/}.}

\subsection{Case Study: Call the Real-World APIs}
\label{subsec:case_study}
Although our datasets are partially synthesized by GPT-4, in fact, this would not affect the usage of real-world tools based on our proposed framework. To demonstrate this, we provide some practical examples of calling real-world APIs in Figure~\ref{fig:case_study}.
It can be seen that our model could generalize to real-world tools well, simply by describing the new tool following our format in the candidate toolset. More examples are shown in Appendix~\ref{app:experiment}.

\section{Conclusion}
In this paper, we propose a decision-aware and generalizable tool-usage framework (DEER) to enhance LLMs' decision-making awareness when addressing diverse user's queries. Meanwhile, we propose a mixture of tool sampling strategy to further improve the generalization of LLMs over unseen tools. Extensive experiments demonstrate the effectiveness of our proposed methods.

\section*{Limitations}
In this work, we propose a decision-aware and generalizable tool-usage framework for large language models and demonstrate its advantages compared to the baselines through extensive experiments. However, there are some limitations so far:
(\romannumeral 1) Considering that collecting numerous tool APIs from the real-world is labor-intensive and time-consuming and in this paper we focus on exploring the LLMs' decision-making awareness and the generalizability of tool-usage, therefore, it is feasible to synthesize diverse tool APIs through GPT-4. In addition, we also demonstrate in Section~\ref{subsec:case_study} that this tool-usage framework we proposed can directly generalize the real-world tool APIs.
(\romannumeral 2) Limited by computational resources, we do not scale our method on larger LLMs (e.g., LLaMA2-70B). Nonetheless, the improvements are observed from LLaMA-7B to LLaMA-13B, thus we believe that better results can be obtained with larger models.

\section*{Ethics Statement}
Our proposed tool-usage framework can effectively reduce the resource consumption of LLMs when manipulating diverse tools. It could be helpful in decreasing carbon emissions, thus making the application and deployment of LLMs more environmentally friendly and sustainable. In addition, all models and datasets used in our experiments are publicly available and have not been reported to carry social bias against any sensitive attributes, and the method we propose would not explicitly introduce new negative societal impacts.

\bibliography{anthology,custom}
\bibliographystyle{acl_natbib}

\appendix
\clearpage

\section{Appendix}

\subsection{More Details of Implementation}
\label{app:implement}
In our experiments, we employ the transformers library from HuggingFace to fine-tune our models, where the default configurations of Trainer (e.g., AdamW optimizer) are used in the training process. Meanwhile, we deploy the LoRA-style parameter-efficient fine-tuning by using HuggingFace PEFT library. In addition, we also leverage Flash Attention \cite{dao2022flashattention} and DeepSpeed\footnote{\url{https://github.com/microsoft/DeepSpeed}} technologies to reduce the requirement of GPU memory in our experiments. Note that all experiments are conducted on $8$ NVIDIA V100 GPUs (32G).
In addition, we describe the detailed algorithmic process of sampling strategies in Algorithm~\ref{alg:sampling}.

\subsection{More Experimental Results}
\label{app:experiment}
\begin{figure*}[!h] 
    \centering
    \subfigure [Bottleneck Size] {
        \includegraphics[width=0.4\textwidth]{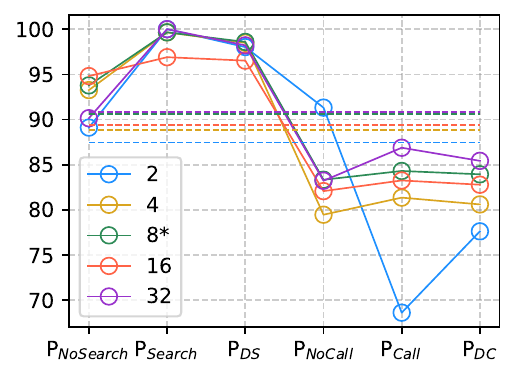}
        \label{subfig:abla_training_loraR}
    }
    \subfigure [Batch Size] {
        \includegraphics[width=0.4\textwidth]{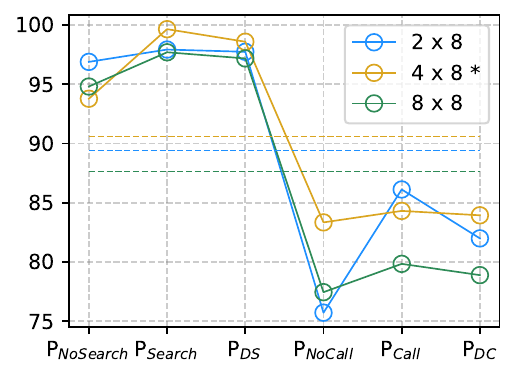}
        \label{subfig:abla_training_bsz}
    } \\
    \subfigure [Learning Rate] {
        \includegraphics[width=0.4\textwidth]{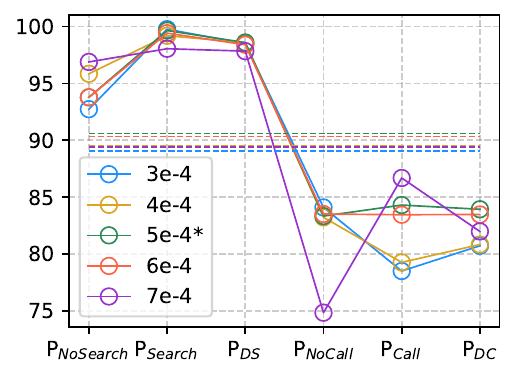}
        \label{subfig:abla_training_lr}
    }
    \subfigure [Number of Epochs] {
        \includegraphics[width=0.4\textwidth]{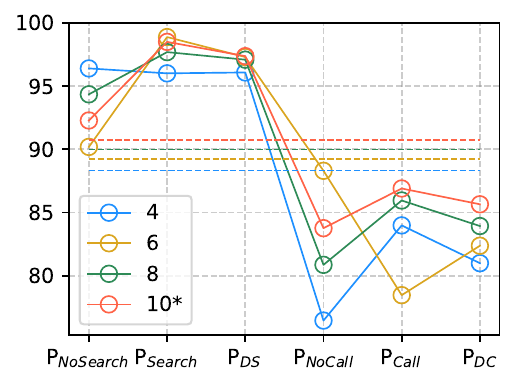}
        \label{subfig:abla_training_epoch}
    }
    \caption{More experiments on hyperparameters during training. * denotes the setting we used in our experiments.}
    \label{fig:abla_training}
\end{figure*}

\begin{figure*}[!h]
    \centering
    \includegraphics[width=1\textwidth]{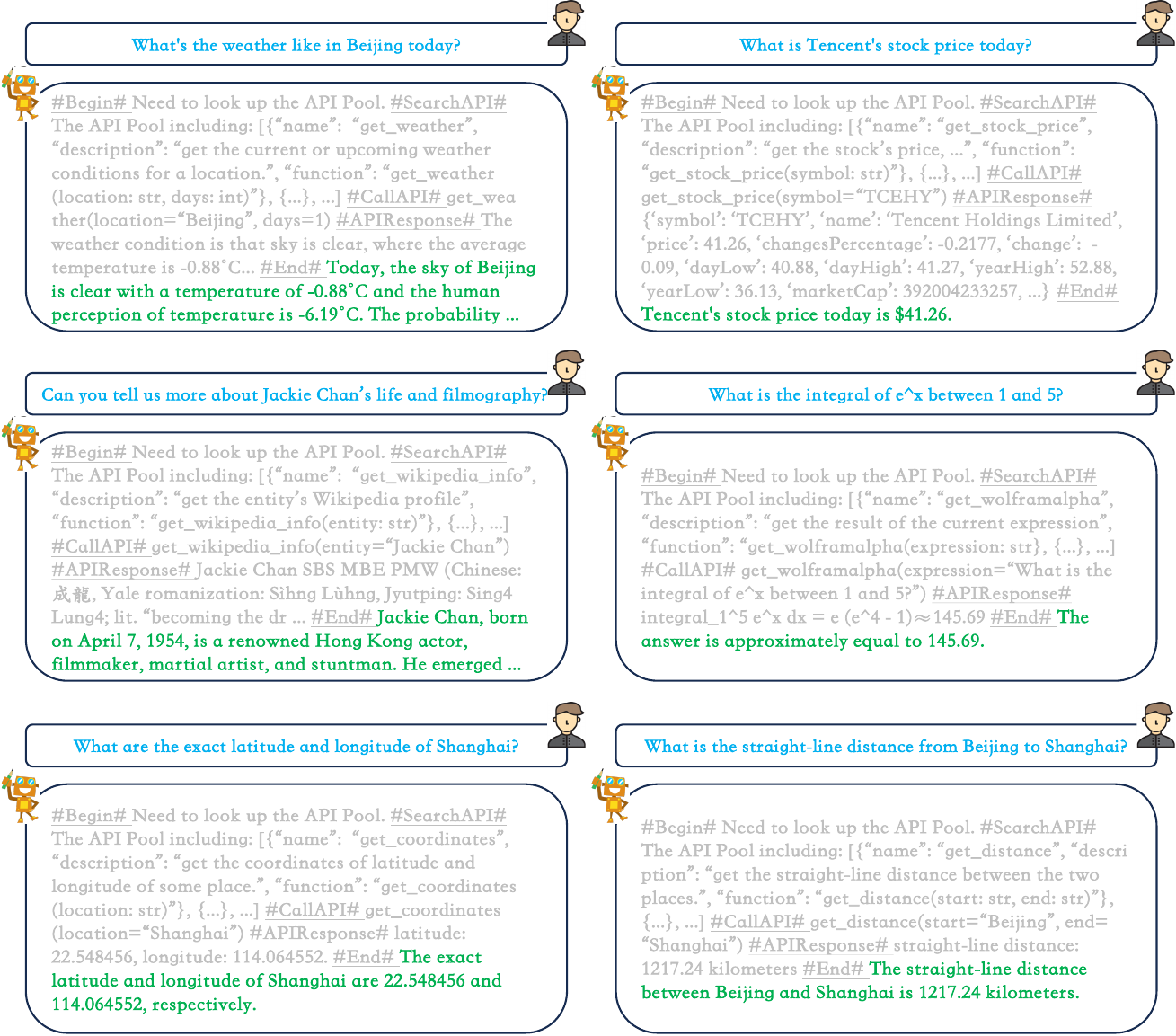}
    \caption{More examples of calling the real-world APIs (e.g., $\texttt{get\_weather}$, $\texttt{get\_stock\_price}$, $\texttt{get\_wikipedia\_info}$, $\texttt{get\_wolframalpha}$, $\texttt{get\_coordinates}$, $\texttt{get\_distance}$)\protect\footnotemark, where the \textcolor{gray}{gray} is an intermediate reasoning step (invisible for the user), the \textcolor{cyan!20!green}{green} is final response to the user's \textcolor{cyan}{query}.}
    \label{fig:extra_case_study}
\end{figure*}

\begin{figure*}[!h]
    \centering
    \includegraphics[width=1\textwidth]{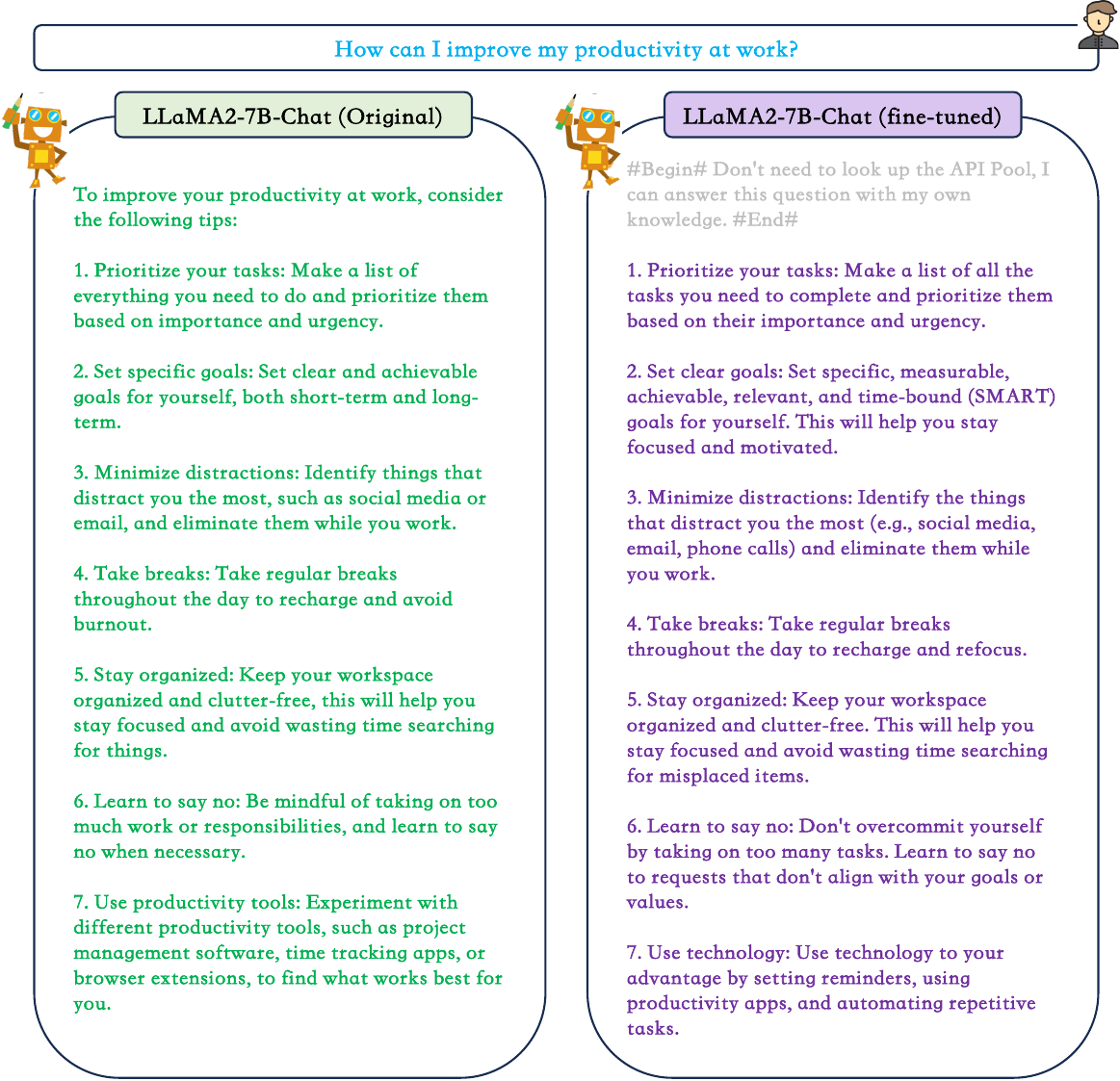}
    \caption{An example of comparing general conversational capability based on LLaMA2-7B.}
    \label{fig:app_general_7B}
\end{figure*}

\begin{figure*}[!h]
    \centering
    \includegraphics[width=1\textwidth]{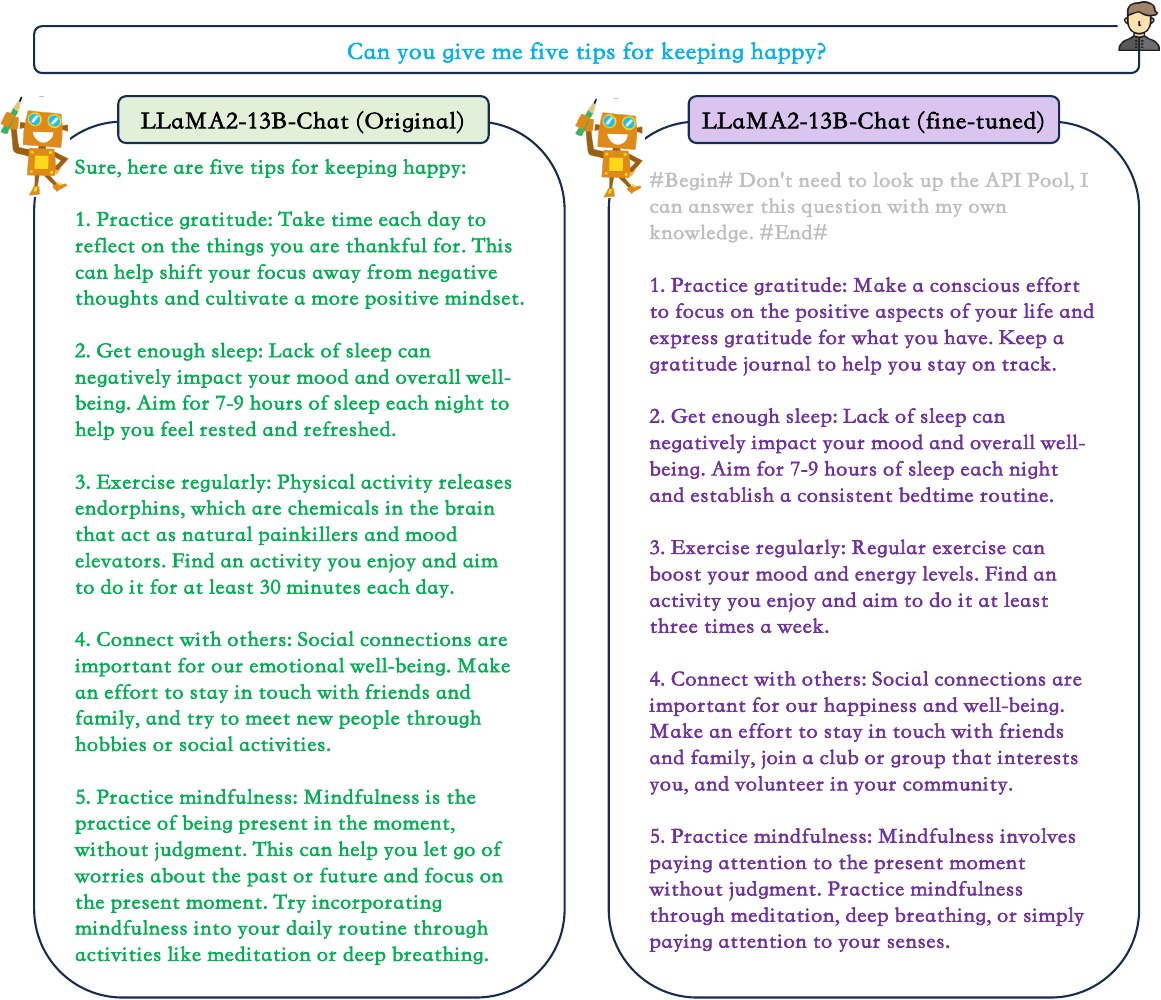}
    \caption{An example of comparing general conversational capability based on LLaMA2-13B.}
    \label{fig:app_general_13B}
\end{figure*}

To determine the optimal setting of hyperparameters during training, we conduct extensive experiments on the bottleneck size of LoRA, the batch size, the learning rate, and the number of epochs. The results are shown in Figure~\ref{fig:abla_training}.
Here, we also provide more practical examples in Figure~\ref{fig:extra_case_study} to demonstrate that our tool-usage framework can seamlessly transfer to real-world tools.
In addition, more comparisons of general conversational capability are presented in Figure~\ref{fig:app_general_7B} and Figure~\ref{fig:app_general_13B}.

\subsection{More Details of Templates}
\label{app:template}
\begin{figure*}[!h]
    \centering
    \includegraphics[width=1\textwidth]{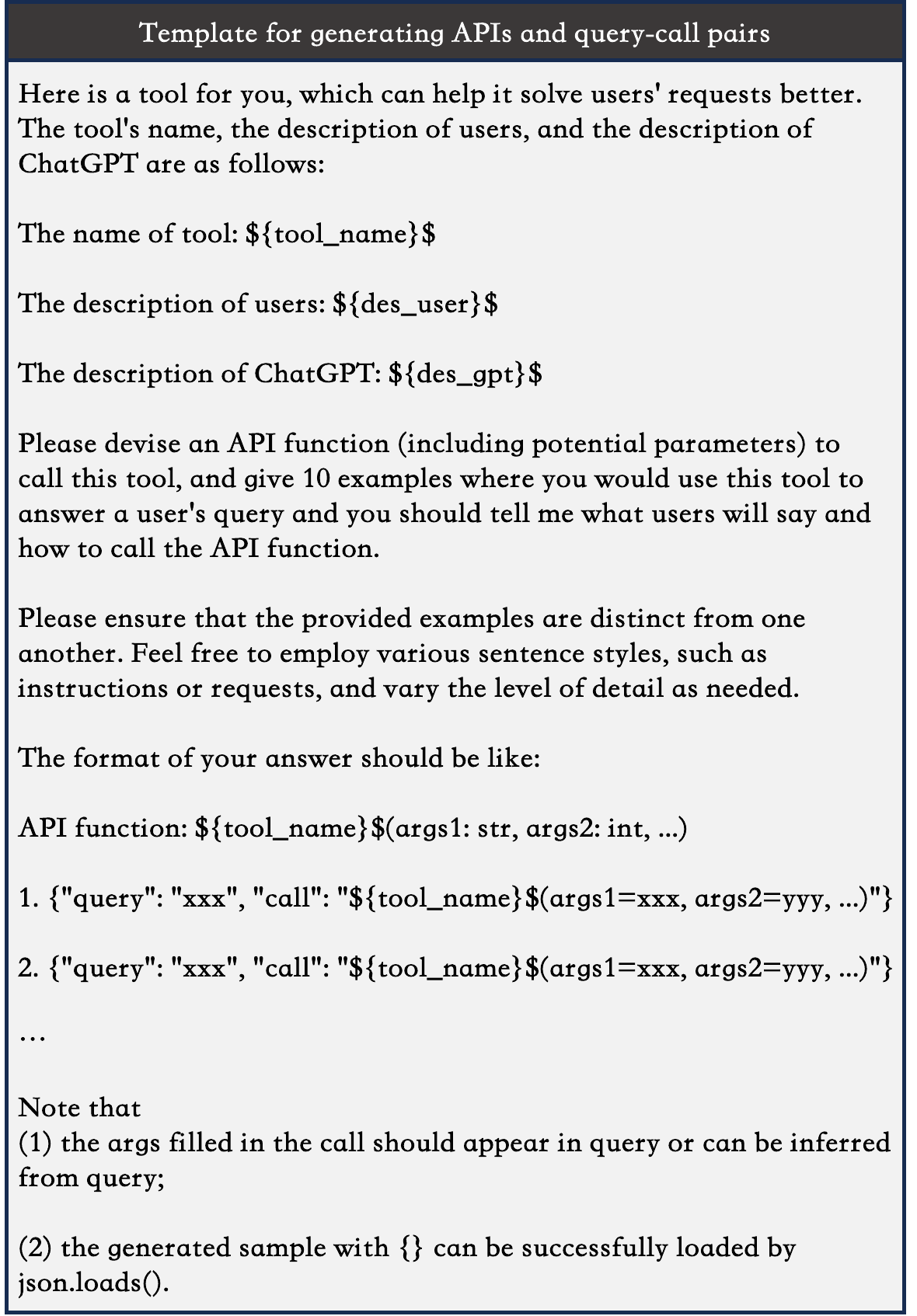}
    \caption{Template of GPT-4 for generating the tool's API and the corresponding query-call pairs.}
    \label{fig:app_tool_gen}
\end{figure*}

\begin{figure*}[!h]
    \centering
    \includegraphics[width=1\textwidth]{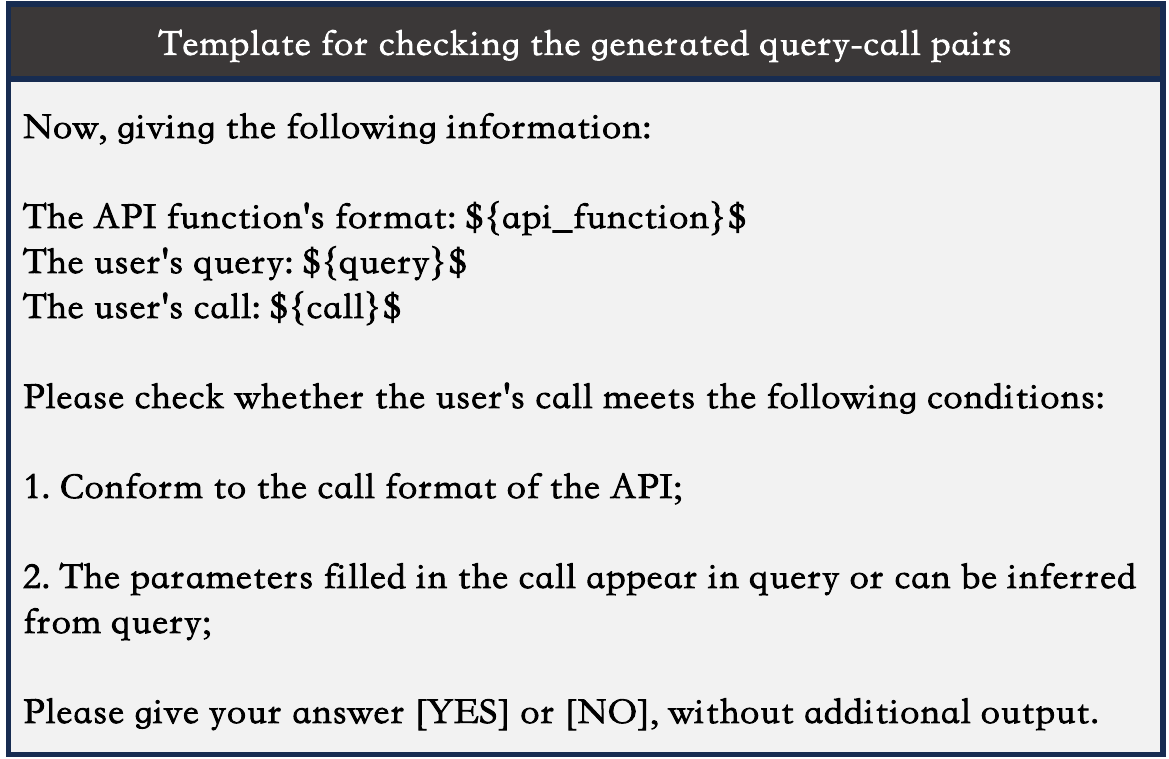}
    \caption{Template of GPT-4 for checking whether the generated query-call pairs make sense.}
    \label{fig:app_check}
\end{figure*}

\begin{figure*}[!h]
    \centering
    \includegraphics[width=1\textwidth]{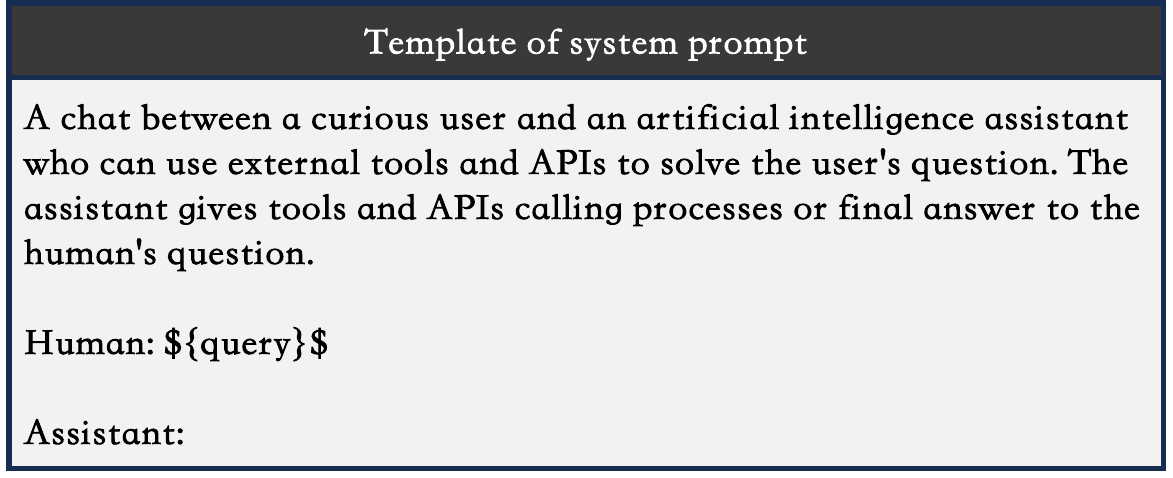}
    \caption{Template of system prompt in our all experiments (including training and inference).}
    \label{fig:app_system}
\end{figure*}

\begin{figure*}[!h]
    \centering
    \includegraphics[width=1\textwidth]{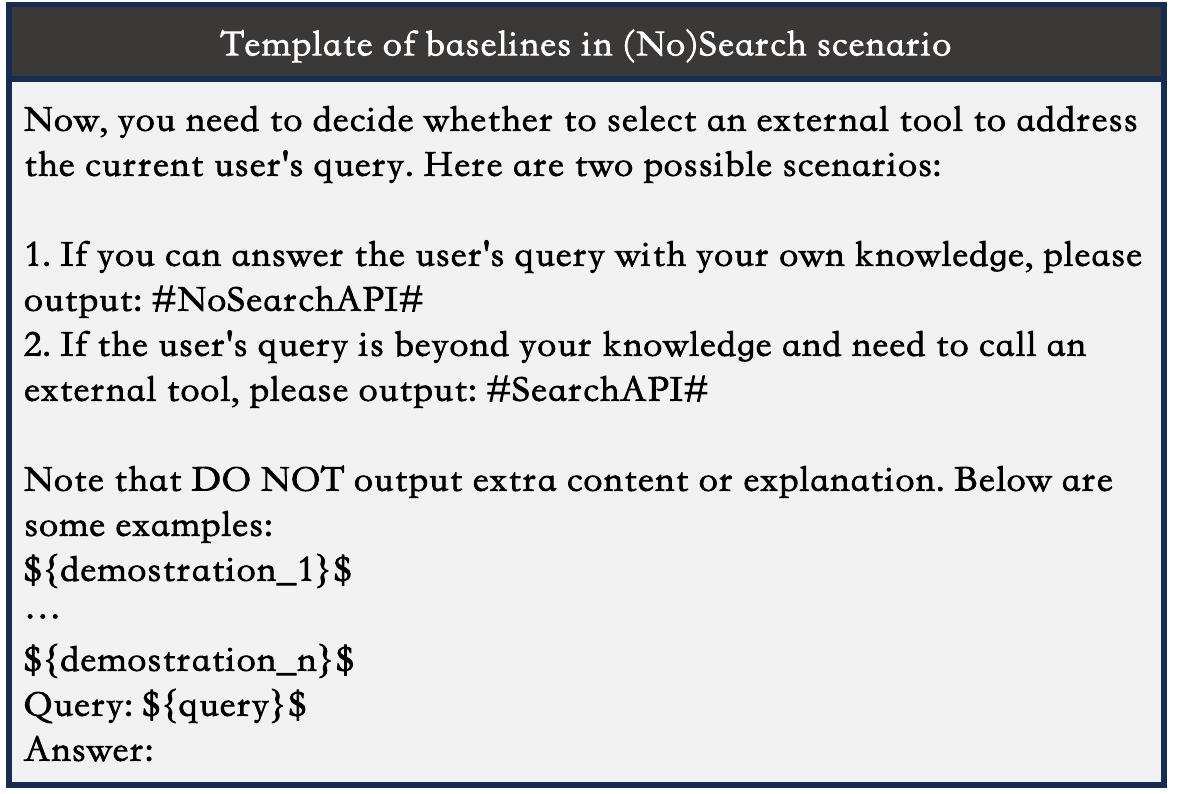}
    \caption{Template of baselines (i.e., ChatGPT, GPT-4, LLaMA2-13B-Chat) in \texttt{Decision-Search} scenario.}
    \label{fig:app_baseline_case1}
\end{figure*}

\begin{figure*}[!h]
    \centering
    \includegraphics[width=1\textwidth]{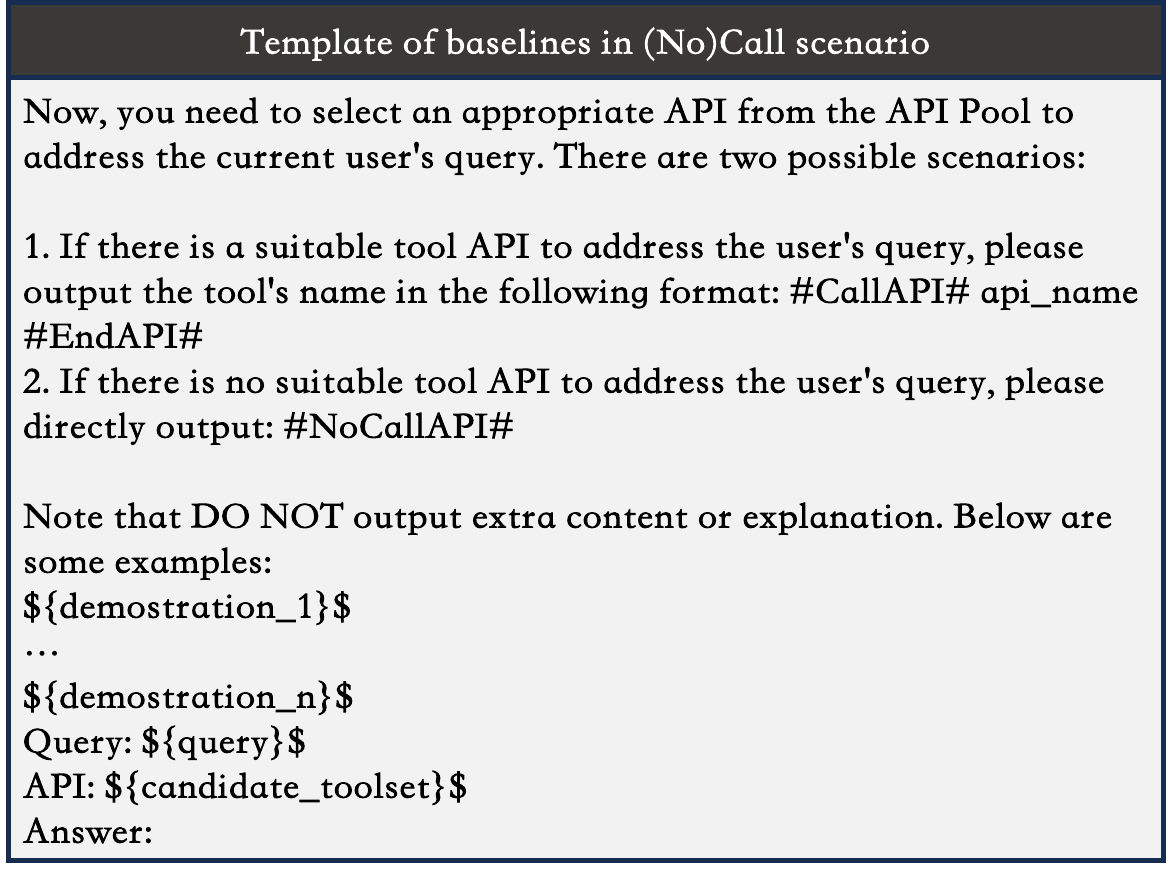}
    \caption{Template of baselines (i.e., ChatGPT, GPT-4, LLaMA2-13B-Chat) in \texttt{Decision-Call} scenario.}
    \label{fig:app_baseline_case2}
\end{figure*}

\begin{figure*}[!h]
    \centering
    \includegraphics[width=1\textwidth]{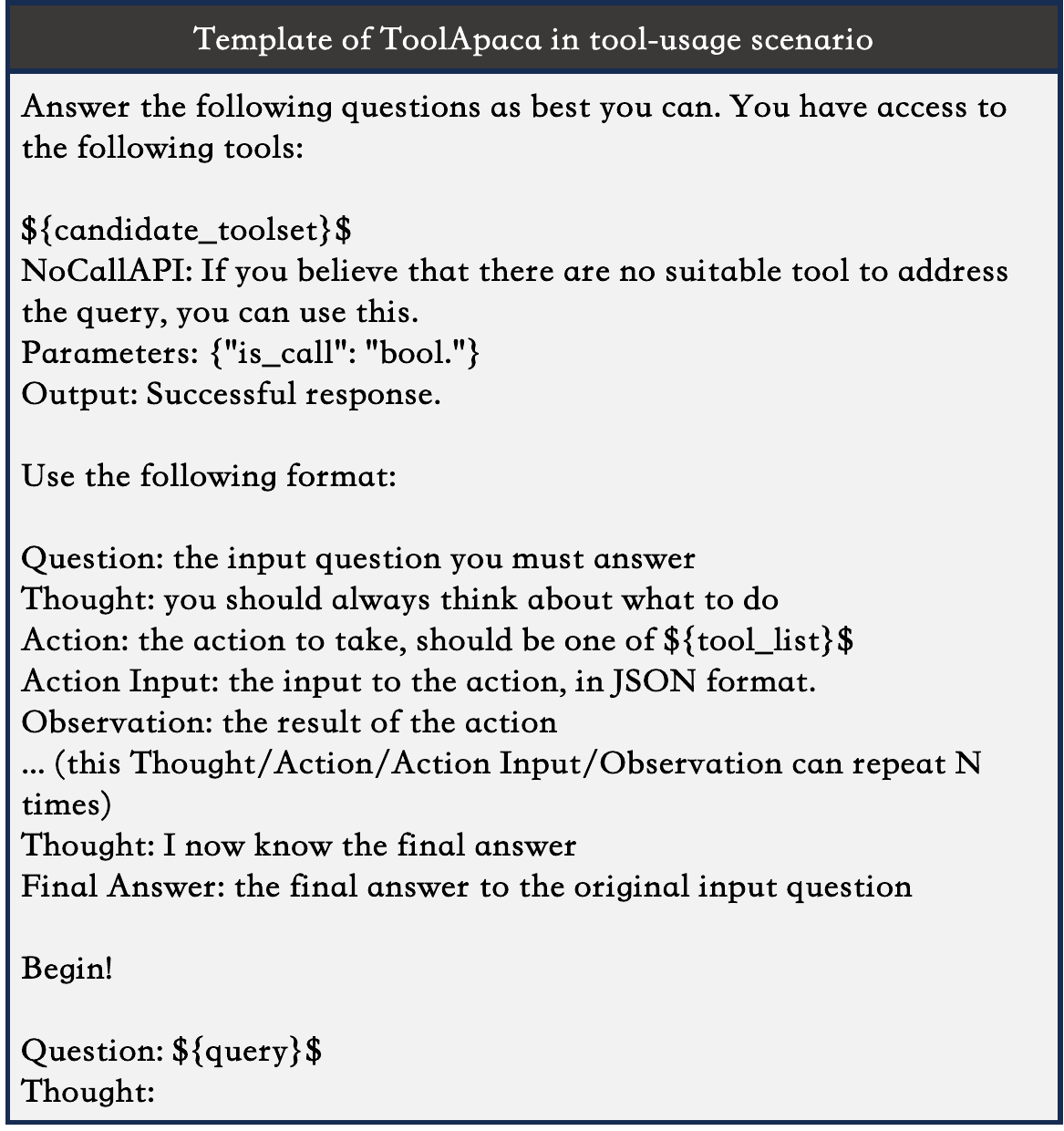}
    \caption{Template of ToolAlpaca for evaluating the generalization on unseen tools.}
    \label{fig:app_toolalpaca}
\end{figure*}

\begin{figure*}[!h]
    \centering
    \includegraphics[width=0.95\textwidth]{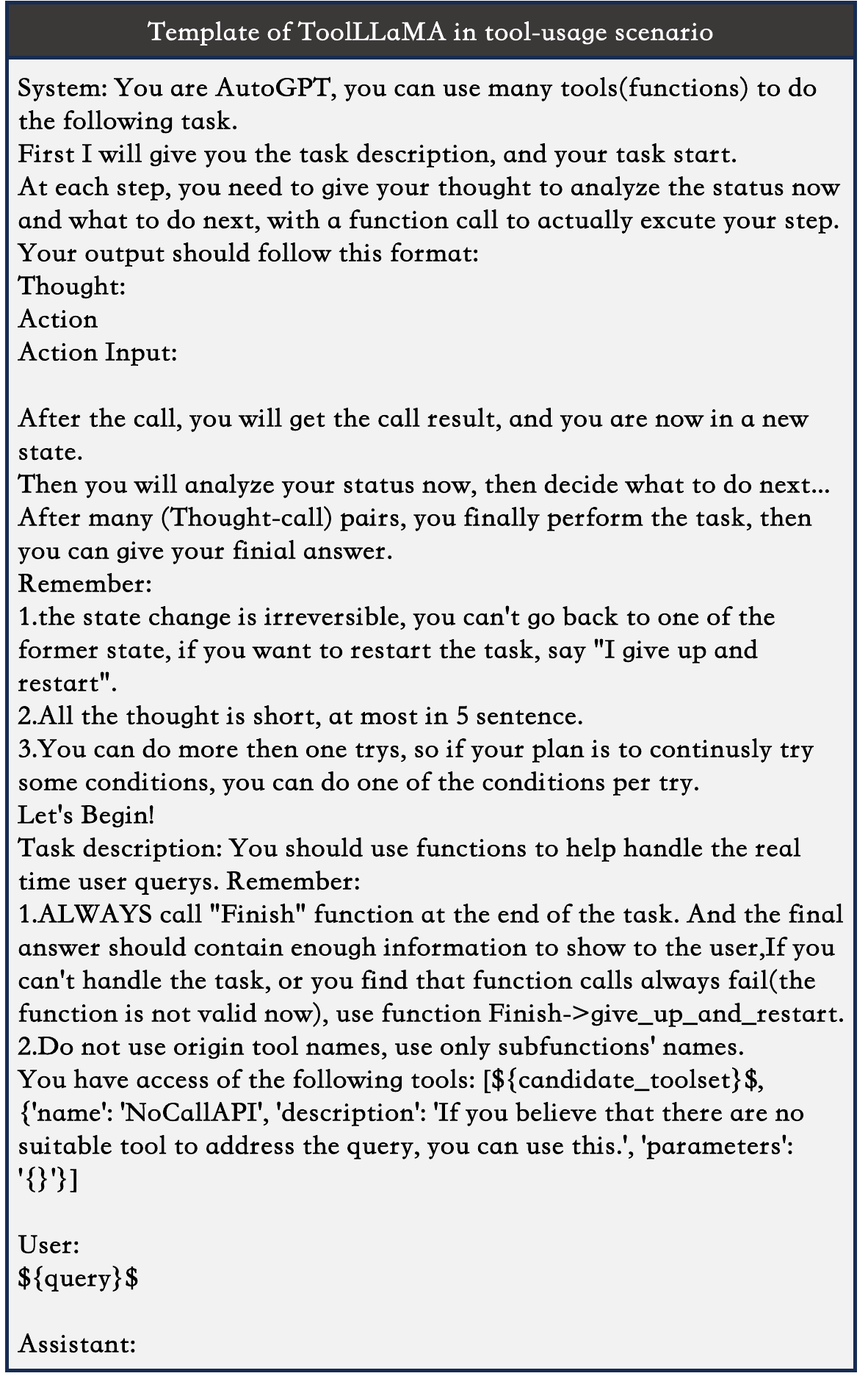}
    \caption{Template of ToolLLaMA for evaluating the generalization on unseen tools.}
    \label{fig:app_toolllama}
\end{figure*}

In this section, we present all the templates used in our experiments. In the data generation process, we first prompt GPT-4 to generate tool's API and the corresponding query-call pairs using the template in Figure~\ref{fig:app_tool_gen}, and then check whether the generated query-call pair is reasonable by using the template in Figure~\ref{fig:app_check}. During the training and inference, we employ the template in Figure~\ref{fig:app_system} as the system template.
Besides, to evaluate the decision-making awareness of baselines (i.e., ChatGPT, GPT-4, LLaMA2-13B-Chat), we adopt the template in Figure~\ref{fig:app_baseline_case1} with some demonstrations. Similarly, the template in Figure~\ref{fig:app_baseline_case2} is utilized in the \texttt{Decision-Call}.
Finally, for the inference of ToolAlpaca and ToolLLaMA on unseen tools, we use their original templates except for introducing a new tool (i.e., ``NoCallAPI'') as a flag for not calling the tool, as shown in Figure~\ref{fig:app_toolalpaca} and Figure~\ref{fig:app_toolllama}.
Note that \$\{...\}\$ in templates refers to the variable that will be populated in the implementation.

\begin{algorithm}[!t]
    \caption{Sampling Strategies for Constructing the Candidate Toolset}
    \textbf{Input:} The set of all tools $\mathcal{T} = \{t_{i}\}_{i=1}^{N}$, the user's query $q$ and corresponding tool $t_{q}$ (with its description $d$). The number of candidate tools and clusters are $k$ and $m$, respectively. The sampling strategy is $\texttt{mode} \in \{\texttt{random}, \texttt{inter-class}, \texttt{intra-class}\}$. \\
    \textbf{Output:} The candidate toolset $\mathcal{T}_{c}$.
    
    \begin{algorithmic}[1] 
        \STATE \# \textbf{Step-1: Embedding}
        \FOR {$t_{i} \in \mathcal{T}$}
            \STATE $e(t_{i}) = \mathrm{embedding}(t_{i}(d))$
        \ENDFOR
        \STATE The tool's vectorized set $\mathcal{E} = \{e(t_{i})\}_{i=1}^{N}$.
        \STATE \# \textbf{Step-2: Clustering}
        \STATE $\Omega \gets \text{K-means}(\mathcal{E}, m)$
        \STATE where $\Omega = \bigcup_{i=1}^{m} \big( C_{i} = \{t^{(i)}_{1}, t^{(i)}_{2}, \cdots, t^{(i)}_{|c_{i}|} \} \big)$.
        \STATE \# \textbf{Step-3: Sampling}
        \IF {$\texttt{mode} == \texttt{random}$}
            \FOR {$1 \le i \le k$}
                \STATE $t_{i} \leftarrow \mathcal{T}$ \# randomly sampling
            \ENDFOR
            \STATE $\mathcal{T}_{c} = \{t_{i}\}_{i=1}^{k}$
        \ELSIF {$\texttt{mode} == \texttt{inter-class}$}
            \STATE $\Omega_{c} \gets \Omega$ \# randomly sampling
            \STATE where $\Omega_{c} = \{C_{i}\}_{i=1}^{k}$.
            \FOR {$C_{i} \in \Omega_{c}$}
                \STATE $t_{i} \gets C_{i}$ \# randomly sampling
            \ENDFOR
            \STATE $\mathcal{T}_{c} = \{t_{i}\}_{i=1}^{k}$
        \ELSIF {$\texttt{mode} == \texttt{intra-class}$}
            \STATE $C^{*} \gets \Omega$
            \STATE where we ensure that $t_{q} \in C^{*}$.
            \FOR {$1 \le i \le k$}
                \STATE $t_{i} \leftarrow C^{*}$ \# randomly sampling
            \ENDFOR
            \STATE $\mathcal{T}_{c} = \{t_{i}\}_{i=1}^{k}$
        \ENDIF
    \end{algorithmic}
    \label{alg:sampling}
\end{algorithm}

\footnotetext{Weather API: \url{https://openweathermap.org/}; Stock API: \url{https://site.financialmodelingprep.com/}; Wikipedia API: \url{https://en.wikipedia.org/w/index.php?search={}}; Wolframalpha API: \url{https://products.wolframalpha.com/api/}; Coordinates \& Distance API: \url{https://lbsyun.baidu.com/faq/api?title=webapi}.}

\end{document}